%% file: main.tex
\documentclass[letterpaper]{article} 
\usepackage{aaai2027}  

\usepackage{eso-pic}
\AddToShipoutPictureFG{%
  \AtPageLowerLeft{%
    \put(\LenToUnit{0.5\paperwidth},\LenToUnit{0.35in}){%
      \makebox[0pt][c]{%
        \parbox{0.85\paperwidth}{\centering\scriptsize\itshape
          This material is provided for informational purposes only and is not
          intended to be investment advice or a recommendation to take any
          particular investment action.}}}}}
          
\usepackage[hyphens]{url}  
\usepackage{graphicx} 
\usepackage{natbib}  
\usepackage{caption} 
\usepackage{algorithm}
\usepackage{algorithmic}

\usepackage{newfloat}
\usepackage{listings}
\DeclareCaptionStyle{ruled}{labelfont=normalfont,labelsep=colon,strut=off} 
\floatstyle{ruled}
\newfloat{listing}{tb}{lst}{}
\floatname{listing}{Listing}

\usepackage{booktabs}
\nocopyright

\usepackage{amsmath,amssymb}

\usepackage{amsfonts}
\usepackage{bm}             
\usepackage{booktabs}      
\usepackage{multirow}      
\usepackage{algorithm}     
\usepackage{algorithmic}   
\usepackage{xcolor}        

\usepackage{tcolorbox}
\tcbuselibrary{skins,breakable}
\newtcolorbox{qualex}[1]{
  colback=gray!5, colframe=gray!55!black, boxrule=0.5pt,
  breakable, arc=2pt, left=6pt, right=6pt, top=4pt, bottom=4pt,
  title=#1, fonttitle=\bfseries\footnotesize
}

\newtheorem{assumption}{Assumption}

\newtheorem{theorem}{Theorem}
\newtheorem{corollary}{Corollary}
\newtheorem{remark}{Remark}
\newcommand{\R}{\mathbb{R}}
\newcommand{\E}{\mathbb{E}}
\newcommand{\Var}{\mathrm{Var}}
\newcommand{\Ind}{\mathbb{1}}
\newcommand{\inner}[2]{\langle #1,\, #2 \rangle}
\newcommand{\norm}[1]{\lVert #1 \rVert}
\newcommand{\clip}{\operatorname{clip}}

\newcommand{\Cov}{\operatorname{Cov}}
\newcommand{\sae}{\operatorname{SAE}}
\newcommand{\sgn}{\operatorname{sign}}

\title{Right Token, Right Time: From Per-Token Gating to Trajectory-Aware Activation Steering}

\author{
    Nima Eshraghi\textsuperscript{\rm 1},
    Lovedeep Gondara\textsuperscript{\rm 1},
    Sagarika Suresh\textsuperscript{\rm 1},
    Xiaotong (Tone) Xu\textsuperscript{\rm 2},\\
    Jithin Pradeep\textsuperscript{\rm 1},
    Fanny Chevalier\textsuperscript{\rm 2}
}
\affiliations{
    \textsuperscript{\rm 1}The Vanguard Group, Inc.\\
    \textsuperscript{\rm 2}University of Toronto\\
    \{nima\_eshraghi, lovedeep\_gondara, sagarika\_thimmanayakanapalya, jithin\_pradeep\}@vanguard.com,\\
    \{tonexu, fanny\}@cs.toronto.edu\\
}

\begin{document}

\maketitle

\begin{abstract}
Activation steering via sparse autoencoders (SAEs) enables behavioral control of
large language models without task-specific fine-tuning. Yet, standard methods apply the steering signal at every generated token, displacing representations from the distribution the model was trained on and eroding generation quality. We show that \emph{when} along a generation the intervention is applied is a decisive axis of control. We cast steering as a per-token allocation problem under a fixed budget of steerable tokens, so that methods differ only in which tokens they select, making allocation the sole variable. Within this framework we introduce two gates. The first, Contrastive Feature Projection (CFP), is memoryless and selects the tokens least aligned with the target behavior through a sparse, interpretable feature projection. The second, Closed-Loop Adaptive Bernoulli Steering (CLABS), is trajectory aware. It treats allocation as closed-loop feedback control, tracking the alignment a generation has accumulated so far and spending budget where the trajectory falls behind. On a linear steering-response model we derive an explicit gain region that guarantees closed-loop stability of CLABS and show that its integral term reduces the steady-state tracking offset. Empirically, CFP and CLABS achieve the largest behavioral shift among budgeted gates at matched budgets, with CLABS strongest under the tight budgets where allocation matters most, while best preserving fluency.
\end{abstract}

\section{Introduction}
Large language models (LLMs) increasingly require \emph{controllable}
behavior at inference time, whether suppressing toxicity, adjusting affect, or
enforcing stylistic constraints, without the cost of retraining. Activation
steering, which adds a fixed direction to a model's hidden states, has emerged
as a lightweight alternative to fine-tuning for such control
\cite{turner2023activation,li2023inference,rimsky2024caa,zou2023representation,eshraghi2026sts}.
Sparse autoencoders (SAEs) sharpen this approach by decomposing hidden states
into interpretable, often monosemantic features
\cite{cunningham2023sparse,bricken2023monosemanticity,templeton2024scaling,lieberum2024gemmascope},
which supply behavior-selective directions along which to steer.

Steering acts directly in representation space, and that directness makes it
easy to adopt. Most methods apply the intervention \emph{uniformly}, pushing
every generated token by the same vector with the same magnitude
\citep{turner2023activation,zou2023representation,rimsky2024caa}. Many tokens,
however, already agree with the target behavior, so steering them adds little
behavioral change while still displacing the residual stream from the
distribution the model was trained on \citep{li2023inference,lee2024cast}. The
displacement accumulates across the sequence, and uniform steering therefore
trades away fluency as behavioral strength grows, a degradation reported
whenever intervention magnitude is pushed high
\citep{li2023inference,rimsky2024caa,templeton2024scaling} and one we
reproduce. If a smaller set of tokens carries most of the behavioral signal,
effective control hinges on \emph{when} to intervene, a question uniform
steering never asks.

Recent work has begun to ask it from several directions. Adaptive magnitude
control regulates \emph{how much} to steer with a feedback controller across
layers, but still intervenes at every position \cite{nguyen2026pid}. Stochastic
sparse steering meets a budget by flipping an independent coin per token
\cite{eshraghi2026sts}, so its selection is unrelated to need. Conditional
steering intervenes only where a trigger fires, for instance on tokens that
activate SAE toxicity features \cite{goyal2025breaking}, at high-entropy
positions selected with a learned value model and multi-sample search
\cite{hu2026sia}, or when tracked states indicate that steering is necessary,
with backtracking to regenerate offending spans \cite{cheng2025fasb}. None of
these, however, combines two properties that a fixed-budget gate needs. The
first is calibration to an explicit intervention budget, so that methods can be
compared at equal cost. The second is a firing rule that responds to the
alignment the generation has \emph{accumulated} so far, rather than to the
current token alone, without auxiliary rollouts.

We study this question within one framework, \emph{sparse} token-level
steering, in which a gate decides whether to intervene subject to a
\emph{budget} $p$, the fraction of tokens that may be steered. Treating
steering as \emph{allocation}, that is, which tokens to steer rather than how
hard to push, turns every comparison into a controlled ablation, since at a
fixed budget methods differ only in which tokens they select. We instantiate
the framework with two gates. \emph{Contrastive Feature Projection} (CFP) reads
each token's instantaneous alignment along a sparse, interpretable feature
direction and spends the budget on the least-aligned positions, at the cost of
one dot product per token. CFP is informed but memoryless, so it cannot
distinguish a momentary dip inside an on-target span from a sustained drift
away from the target. \emph{Closed-Loop Adaptive Bernoulli Steering} (CLABS),
our central method, adds this memory. It treats allocation as feedback control,
setting a budget-calibrated steering probability from the current misalignment
and a leaky integral of recent misalignment, so that sustained drift draws more
budget than isolated dips. CLABS runs within one decoding trajectory, without
auxiliary rollouts, backtracking, or a learned scorer.
Our contributions are as follows.
\begin{itemize}
  \item We cast token-level activation steering as a \emph{budgeted allocation} problem, in which \emph{when} to spend a fixed intervention budget is the single decision variable, and introduce two efficient activation-space gates, the memoryless CFP and the trajectory-aware CLABS.

  \item We analyze CLABS as a proportional--integral controller with a leaky integral term. Under a linear approximation of the steering response, common in feedback analysis \cite{nguyen2026pid}, we derive an explicit gain region that guarantees closed-loop stability and bounded tracking error, and show that integral action reduces, but does not remove, the steady-state offset. To our knowledge, this is the first stability certificate for a sparse steering gate.

  \item On two base models and three span-level targets, CFP and CLABS achieve the largest behavioral shift among budgeted gates at matched realized budgets, with the widest margin at tight budgets. Among the budgeted gates they best preserve fluency and diversity, and under toxicity reduction they are the only gates preferred to the unsteered model on instruction following at every budget under both judges. A component ablation isolates the contribution of the history term.
\end{itemize}

\section{Related Work}\label{sec:related}
\paragraph{Activation steering.}
Activation steering adds a fixed direction to a model's hidden states at
inference time, controlling behavior without weight updates
\citep{subramani2022latent,turner2023activation,rimsky2024caa,li2023inference,zou2023representation}.
Directions have been derived from class-conditional mean differences
\citep{marks2024geometry}, in-context demonstrations \citep{liu2024icv}, weight
arithmetic \citep{fierro2026weight}, and expert (de)activation in
Mixture-of-Experts models \citep{fayyaz2026steermoe}. Sparse autoencoders
decompose the residual stream into sparse, near-monosemantic features
\citep{cunningham2023sparse,bricken2023monosemanticity,templeton2024scaling,lieberum2024gemmascope},
whose decoder rows provide interpretable steering directions
\citep{soo2025fgaa}. AxBench \citep{wu2025axbench} shows that the choice of
direction matters and that simple baselines can rival SAE-based ones, which is
why we hold the direction fixed across all methods and vary only allocation.
Steering now spans detoxification, truthfulness, hallucination reduction,
style, safety, and instruction following
\citep{turner2023activation,li2023inference,yang2025casal,delangis2024multireward,stolfo2025instruct},
complementing decoding-time detoxification controllers
\citep{dathathri2020plug,liu2021dexperts} evaluated on RealToxicityPrompts
\citep{gehman2020realtoxicityprompts}. Most of these methods apply the
direction \emph{at every generated token}, and this dense intervention degrades
fluency as behavioral shift grows
\citep{turner2023activation,rimsky2024caa,wu2025axbench}.
Recent work accordingly evaluates steering on the trade-off between behavioral
strength and quality rather than on either alone
\citep{ferrando2025dynamically}, a view we adopt by reporting efficacy and
quality together at matched budgets.

\paragraph{Sparse, conditional, and adaptive steering.}
Recent work relaxes the every-token assumption along different axes. AUSteer
\citep{feng2026austeer} sparsifies \emph{which dimensions} are edited but still
fires at every position, and is orthogonal to the temporal sparsity we study.
Stochastic Token Steering \citep[STS;][]{eshraghi2026sts} gates each token with
an independent $\mathrm{Bernoulli}(p)$ draw, which meets a budget but selects
positions independently of the state. Conditional methods intervene only where
a trigger fires. Breaking Bad Tokens \citep{goyal2025breaking} applies
SAE-derived interventions at tokens that activate toxicity features, a
threshold rule that corresponds to CFP in our framework once calibrated to a
budget. Sparse Inference-time Alignment \citep[SIA;][]{hu2026sia} steers
high-entropy junctions in logit space through a learned, reward-distilled value
model coupled to Best-of-$N$ search. FASB \citep{cheng2025fasb} tracks internal
states over generated tokens to decide whether and how strongly to steer, and
backtracks to regenerate tokens when steering is deemed necessary. These
methods are governed by thresholds, rewards, or regeneration rather than by an
intervention budget, and their firing decisions do not feed back on the
alignment accumulated along a single trajectory. Where a single-pass selection
signal can be isolated, as for the feature trigger and the entropy criterion,
we compare against it as a budget-calibrated gate rather than against the
complete system. PID Steering \citep{nguyen2026pid} also draws on classical
control, but regulates the steering \emph{magnitude} at every token with an
integral term accumulated across layers and admits no budget. It acts on a
complementary axis, since a gate decides \emph{when} to steer and a magnitude
controller \emph{how much}, and the two compose. CLABS holds the magnitude
fixed and applies PI feedback \emph{along the token sequence} to set a
budget-calibrated \emph{gating probability}. To our knowledge, it is the first
gate that allocates a fixed steering budget by feedback on accumulated
alignment within one decoding trajectory, without rollouts or backtracking,
and the first sparse steering gate with a closed-loop stability analysis.
\section{Method} \label{sec:method}
We present two gates that spend the same budget in different ways. Both read a
single alignment signal, the projection of the residual stream onto a steering
direction built from sparse autoencoder features. \textbf{CFP} is
\emph{memoryless} and steers the least-aligned tokens. \textbf{CLABS} feeds the
same signal into a proportional--integral (PI) controller with a leaky integral
term, so its steering probability also reflects the alignment the generation
has accumulated so far. We then analyze the resulting closed loop. Because both
gates are calibrated to the same budget, allocation is the only difference
between them.

\subsection{Notation and Setup}
\label{sec:notation-setup}
Let $p_\theta$ be an autoregressive language model and $h_t^{\ell}\in\R^d$
the residual-stream activation at a fixed intervention layer $\ell$ and
decoding step $t$. We write $\inner{\cdot}{\cdot}$ for the Euclidean inner
product, $\norm{\cdot}$ for the induced norm, $\mathbb{1}[\cdot]$ for the
indicator, $\sgn(\cdot)$ for the sign, $\mu,\sigma$ for a sample mean and
standard deviation, $\epsilon>0$ for a small constant, and
$\clip(x;a,b)=\min(\max(x,a),b)$.

A token-level steering intervention adds a fixed direction to the residual
stream, gated per token,
\begin{equation}
\tilde h_t^{\ell} = h_t^{\ell} + m_t\,\alpha\,v,
\label{eq:intervention-main}
\end{equation}
where $v\in\R^d$ with $\norm{v}=1$ is the steering direction, $\alpha\in\R$
the strength, and $m_t\in\{0,1\}$ the gate. This separates two control axes,
\emph{allocation}, given by the gate sequence $m_{1:T}$, and \emph{magnitude},
given by $\alpha$. We hold $\alpha$ fixed and study allocation. 
Every method runs at the same \emph{budget} $p\in(0,1)$, the expected fraction of steered tokens.
Differences in downstream behavior are therefore attributable to \emph{which} tokens a method steers.

\subsection{Steering Direction from Sparse Features}
\label{sec:directions}
We derive the direction from a sparse autoencoder (SAE) rather than from raw
activation differences. SAEs decompose the residual stream into sparse,
approximately monosemantic features
\cite{cunningham2023sparse,bricken2023monosemanticity}. A direction built from
a few behavior-selective features is therefore expected to be less entangled
with unrelated attributes than a dense difference-in-means vector, which
matters when the goal is to move one behavior while leaving generation quality
intact.

Let the SAE at layer $\ell$ have encoder $f:\R^d\to\R^F$, decoder
$W_{\mathrm{dec}}\in\R^{F\times d}$, and reconstruction
$\sae(h)=W_{\mathrm{dec}}^\top f(h)$. Let $\mathcal D^+$ be a target corpus
that exhibits the behavior of interest, namely the behavior to be induced for
elicitation or the behavior to be removed for suppression, and let
$\mathcal D^-$ be a neutral corpus. We compute the per-feature differential
\begin{equation}
\Delta_i=\frac{1}{|\mathcal D^+|}\sum_{x\in\mathcal D^+} f_i(x)
-\frac{1}{|\mathcal D^-|}\sum_{x\in\mathcal D^-} f_i(x),
\label{eq:differential}
\end{equation}
where $f_i(x)$ denotes the $i$-th SAE code averaged over the tokens of $x$. We
take the indices $\mathcal I$ of the top-$k$ features by $\Delta_i$ and form
the direction from the corresponding decoder rows,
\begin{equation}
v=\frac{\sum_{i\in\mathcal I} W_{\mathrm{dec},i}}
{\norm{\sum_{i\in\mathcal I} W_{\mathrm{dec},i}}}.
\label{eq:direction}
\end{equation}

Since $v$ points toward $\mathcal D^+$, suppression tasks use $\alpha<0$ to
move activations away from it, while elicitation tasks use $\alpha>0$. We
define a task-agnostic \emph{behavioral probe}
\begin{equation}
u=\sgn(\alpha)\,v,
\label{eq:probe}
\end{equation}
so that a larger $\inner{h_t^{\ell}}{u}$ always indicates that $h_t^{\ell}$
is \emph{more aligned} with the desired behavior, regardless of the task sign.
The probe is a fixed vector, so reading it requires no SAE forward pass at
inference time.

\subsection{CFP: Contrastive Feature Projection}
\label{sec:cfp}
A token already aligned with the target has little to gain from
intervention, whereas a token far from the target is where a fixed-strength
push is expected to change the trajectory most. CFP reads each token's
instantaneous alignment along the probe and spends the budget on the
least-aligned positions,
\begin{equation}
s_t^{\mathrm{CFP}}=\inner{h_t^{\ell}}{u},\qquad
m_t=\mathbb{1}\!\left[s_t^{\mathrm{CFP}}\le\tau\right],
\label{eq:cfp-signal}
\end{equation}
with $\tau$ calibrated to the budget $p$. By sparing tokens that are already
on target, CFP avoids perturbing the residual stream where steering would add
little. CFP is the budget-calibrated form of conditional, feature-triggered
steering. Given $\tau$, each $m_t$ depends only on the current activation, so
CFP cannot tell a momentary dip inside an on-target span from a sustained
drift away from the target. It may therefore keep steering an aligned span, or
stop too early on one that is not. This limitation motivates CLABS.

\subsection{CLABS: Closed-Loop Adaptive Bernoulli Steering}
\label{sec:clabs}
CLABS closes the loop. Instead of thresholding a static signal, it treats
allocation as discrete-time feedback control \citep{astrom1997computer} on the
generation's \emph{running} alignment, so the budget is spent in response to
what has already been generated. We first standardize the projection with
calibration statistics $\mu,\sigma$,
\begin{equation}
\hat s_t=\frac{\inner{h_t^{\ell}}{u}-\mu}{\sigma+\epsilon},
\label{eq:s-standardized}
\end{equation}
which expresses alignment in standard deviations above or below its typical
value. We then choose a \emph{setpoint} $s^\star$ in the same units. It is the
internal alignment level CLABS aims to hold, with $s^\star=0$ targeting the
calibration mean and larger $s^\star$ demanding stronger alignment.  Given $s^\star$, we define the tracking error and its leaky
integral, an exponentially weighted moving average (EWMA),
\begin{align}
e_t &= s^\star-\hat s_t, \label{eq:err}\\
\bar e_t &= (1-\beta)\bar e_{t-1}+\beta e_t,\quad \bar e_0=0, \label{eq:ewma}
\end{align}
where $\beta\in(0,1]$ sets the controller's memory, so that $\bar e_t$
summarizes the deviation accumulated over roughly the last $1/\beta$ tokens.
CLABS then sets a PI steering probability and samples the gate,
\begin{align}
p_t &= \clip\!\big(p_0+\kappa_p e_t+\kappa_i\bar e_t;\,0,1\big), \label{eq:pi}\\
m_t &\sim \mathrm{Bernoulli}(p_t), \label{eq:bernoulli}
\end{align}
with bias $p_0$ and gains $\kappa_p,\kappa_i\ge0$. The proportional term
$\kappa_p e_t$ reacts to the current token's misalignment, the same
information CFP uses. The leaky integral term $\kappa_i\bar e_t$ adds what memoryless gates lack, a record of whether recent tokens have stayed behind the setpoint. The steering probability therefore rises while the trajectory lags and relaxes once it catches up, so a sustained drift draws more budget than an isolated dip. The bias $p_0$ is calibrated so that the realized gate rate matches the target budget $p$, as detailed in Appendix~\ref{app:implementation}. Algorithm~\ref{alg:clabs} summarizes the procedure. Unlike conditional methods that rely on a learned scorer with multi-sample search \citep{hu2026sia} or on backtracking \citep{cheng2025fasb}, CLABS needs none of these. To our knowledge, it is the first gate that sets a budget-calibrated steering probability by feedback on accumulated alignment within a single decoding trajectory.

\begin{algorithm}[t]
\caption{CLABS}
\label{alg:clabs}
\begin{algorithmic}[1]
\REQUIRE probe $u$, direction $v$, strength $\alpha$, setpoint $s^\star$,
         gains $\kappa_p,\kappa_i,\beta$, bias $p_0$, stats $\mu,\sigma$
\STATE $\bar e\gets 0$
\FOR{$t=1,\dots,T$}
  \STATE obtain residual $h_t^{\ell}$ at layer $\ell$
  \STATE $\hat s_t \gets (\inner{h_t^{\ell}}{u}-\mu)/(\sigma+\epsilon)$
  \STATE $e_t\gets s^\star-\hat s_t$;\quad $\bar e\gets (1-\beta)\bar e+\beta e_t$ 
  \STATE $p_t\gets \clip(p_0+\kappa_p e_t+\kappa_i\bar e;\,0,1)$
  \STATE sample $m_t\sim\mathrm{Bernoulli}(p_t)$
  \IF{$m_t=1$}
    \STATE $\tilde h_t^{\ell}\gets h_t^{\ell}+\alpha v$, then apply
           norm preservation and projection clamping
  \ENDIF
  \STATE decode the next token from the (possibly steered) state
\ENDFOR
\end{algorithmic}
\end{algorithm}


\subsection{Theoretical Analysis of CLABS}\label{sec:theory}

A gate driven by the model's own state forms a feedback loop. Steering changes
alignment, alignment changes the next steering decision, and loops of this kind
can oscillate or drift. Before evaluating CLABS empirically, we therefore ask
whether its controller is stable and which gains keep it stable. We answer
both questions on a first-order \emph{steering-response model} of how steering
moves alignment, following standard practice in feedback control of computing
systems~\citep{hellerstein2004feedback,astrom1997computer} and prior
control-theoretic steering~\citep{nguyen2026pid}. The model describes the
interior operating point, where the calibrated budget keeps $p_t$ strictly
inside $(0,1)$. The result is therefore a stability certificate for this
linear approximation, not a worst-case guarantee for the full nonlinear
decoding process. Behavioral efficacy at matched budgets is assessed
separately in Section~\ref{sec:experiments}.

\begin{assumption}[Linear steering response]
\label{assumption:plant}
Let $\mathcal F_{t-1}$ denote the generation history available when $p_t$ is
computed. In the interior regime there exist constants $\phi\in[0,1)$, $g>0$,
and $\beta_0\in\R$ such that
\begin{equation}
\hat s_{t+1}
=\phi\,\hat s_t + (1-\phi)\beta_0 + g\,m_t + \eta_t .
\label{eq:plant}
\end{equation}
Here $\phi$ is the \emph{activation persistence}, measuring how strongly
alignment carries over from one token to the next, $g$ is the \emph{steering
gain}, measuring how much steering a token shifts next-token alignment, and
$\beta_0$ is the \emph{unsteered equilibrium}. The disturbance $\eta_t$ has
zero mean given $\mathcal F_{t-1}$ and conditional variance at most
$\sigma_\eta^2$, and it absorbs token-level fluctuations and neglected
higher-order effects.
\end{assumption}

In the interior regime, where the clip in Eq.~\eqref{eq:pi} is inactive,
combining Eq.~\eqref{eq:plant} with the PI law and the leaky integral of
Eq.~\eqref{eq:ewma} shows that the controller state
$x_t=(e_t,\bar e_t)^\top$ follows the linear recursion
\begin{equation}
\!\!x_{t+1}=A\,x_t+b+w_t,~
A=
\begin{pmatrix}
a & -g\kappa_i\\[2pt]
\beta a & (1-\beta)-\beta g\kappa_i
\end{pmatrix},
\label{eq:state}
\end{equation}
where $a=\phi-g\kappa_p$, $b$ is a constant drift, and $w_t=-r\,\tilde\eta_t$
with $r=(1,\beta)^\top$ is a zero-mean disturbance. The effective noise
$\tilde\eta_t$ accounts for both the model disturbance $\eta_t$ and the
randomness of the Bernoulli gate, and its conditional variance is bounded by a
constant $\bar\sigma^2$. The derivation of this recursion, the explicit form
of $b$, and the bound $\bar\sigma^2$ are given in
Appendix~\ref{appx_thm_proof}. The matrix $A$ couples how the model responds
to steering ($\phi$, $g$) with how the controller reacts ($\kappa_p$,
$\kappa_i$, $\beta$), so stability is a joint property of the model and the
gate.

\begin{theorem}[Stability and tracking]
\label{thm:stability}
Under Assumption~\ref{assumption:plant}, let $\rho(A)$ be the spectral radius
of $A$ in Eq.~\eqref{eq:state}, $I$ the $2\times2$ identity, and $[z]_1$ the
first entry of a vector $z$. A sufficient condition for $\rho(A)<1$ is
\begin{equation}
0\le a<1,\qquad
0\le g\kappa_i<\frac{(2-\beta)(1+a)}{\beta}.
\label{eq:jury-cond}
\end{equation}
Under this condition the following hold.
\begin{enumerate}
\item $\E[x_t]$ converges geometrically, with rate arbitrarily close to
$\rho(A)$, to the \emph{steady-state controller state}
$x_{\mathrm{ss}}=(I-A)^{-1}b$.
\item $\limsup_{t\to\infty}\Var(e_t)\le c(A)\,\bar\sigma^2$, where
$c(A)=\sum_{k\ge0}\big([A^k r]_1\big)^2\ge1$ is a gain-dependent variance
multiplier, finite whenever $\rho(A)<1$.
\item With $e_{\mathrm{ss}}=[x_{\mathrm{ss}}]_1$ the \emph{steady-state
tracking error},
\begin{equation}
\frac{1}{T}\sum_{t=1}^T
\E\!\left[(e_t-e_{\mathrm{ss}})^2\right]
\le O\!\left(\tfrac1T\right)+O(\bar\sigma^2).
\label{eq:rate}
\end{equation}
\end{enumerate}
\end{theorem}
The proof is given in Appendix~\ref{appx_thm_proof}. The theorem reduces
stability of the gated loop to a closed-form condition that involves only
$\phi$, $g$, and the controller gains, so any candidate $(\kappa_p,\kappa_i,\beta)$
can be checked against Eq.~\eqref{eq:jury-cond} given estimates of $\phi$ and
$g$. On the linear model, a choice inside the box guarantees that the mean
controller state converges geometrically, that the asymptotic variance of the
tracking error is at most $c(A)\,\bar\sigma^2$, and that the transient
contribution to the time-averaged squared error vanishes at rate $O(1/T)$.
To our knowledge, no prior sparse steering gate comes with such a
certificate.

\begin{corollary}[Integral action reduces the steady-state offset]
\label{cor:integral}
Under condition~\eqref{eq:jury-cond}, with $p_0$, $\kappa_p$, and the model
constants $\phi,g,\beta_0$ fixed, the deterministic steady-state error is
\begin{equation} \label{eq:ess}
e_{\mathrm{ss}}=
\frac{N}{\,1-\phi+g(\kappa_p+\kappa_i)\,},\quad
N=(1-\phi)(s^\star-\beta_0)-g\,p_0 .
\end{equation}
Hence, as $\kappa_i$ varies within the box, $|e_{\mathrm{ss}}|$ is
non-increasing in $\kappa_i$, and strictly decreasing whenever $N\neq0$.
\end{corollary}
The proof is given in Appendix~\ref{appx_thm_proof}.

\begin{remark}
\label{rem:gains}
The two inequalities in Eq.~\eqref{eq:jury-cond} correspond to the two
controller terms. The first keeps the proportional correction $g\kappa_p$ no
larger than the persistence $\phi$, so the proportional response alone never
flips the sign of the error. The second caps the integral gain by a bound that
grows with the memory length $1/\beta$, so slow integration is easier to
stabilize but reacts more sluggishly. The same box also limits how much
integral action can help. Here $N$ is $(1-\phi)$ times the gap between the
setpoint and the alignment an open-loop gate with bias $p_0$ would reach.
Combining Eq.~\eqref{eq:ess} with Eq.~\eqref{eq:jury-cond} gives, for every
admissible gain at a fixed $p_0$ and $N\neq0$,
\begin{equation}
|e_{\mathrm{ss}}|\;>\;\frac{|N|}{1-a+(2-\beta)(1+a)/\beta}\;>\;0 .
\label{eq:offset-bound}
\end{equation}
Unlike a classical integrator, the leaky integral therefore trades exact
tracking for bounded memory, reducing the offset. The
derivation is given in Appendix~\ref{appx_thm_proof}.
\end{remark}


\section{Experiments}
\label{sec:experiments}
Our experiments test two claims. The first is that at a fixed intervention
budget the choice of \emph{which} tokens are steered materially changes the
behavioral outcome. The second is that trajectory-aware allocation improves on
memoryless allocation, and that the improvement is largest where the budget is
tight. Within each model and target, every method shares the steering
direction, intervention layer, per-token strength, decoding settings, and
post-intervention corrections, and differs only in \emph{when} the direction is
injected. The gating policy is therefore the sole independent variable.

\subsection{Setup}
\label{sec:setup}
We evaluate two open base models that differ in scale and family,
\textsc{Gemma\,2\,2B} \cite{team2024gemma2} and \textsc{Llama\,3.1\,8B}
\cite{dubey2024llama}. Following established practice, we intervene on the
residual stream at a single mid-network layer \cite{labonne2024clamping}, and
read both the steering direction and the gating signals from a pretrained
residual-stream SAE at that layer. We study two control polarities over three
span-level targets. \emph{Suppression} uses \textsc{RealToxicityPrompts}
\cite{gehman2020realtoxicityprompts}, whose continuations we detoxify.
\emph{Elicitation} uses \textsc{GoEmotions} \cite{demszky2020goemotions},
collapsed to the Ekman categories \cite{ekman1992argument}, where we induce
\emph{fear} or \emph{sadness} from neutral prompts. To verify that control does
not come at the expense of competence, we also evaluate instruction following
on \textsc{AlpacaEval} \cite{alpaca_eval,dubois2024alpacafarm}. All results are
averaged over three random seeds. Three lightweight corrections follow
Eq.~\eqref{eq:intervention-main}, namely norm preservation
\cite{turner2023activation}, activation clamping \cite{labonne2024clamping},
and a decoding-stage repetition penalty \cite{keskar2019ctrl}. They are applied
identically to every method, so no gate receives a correction the others do
not. Further implementation details are given in
Appendix~\ref{app:implementation}.

\begin{figure*}[t]
    \centering
    \includegraphics[scale=0.44]{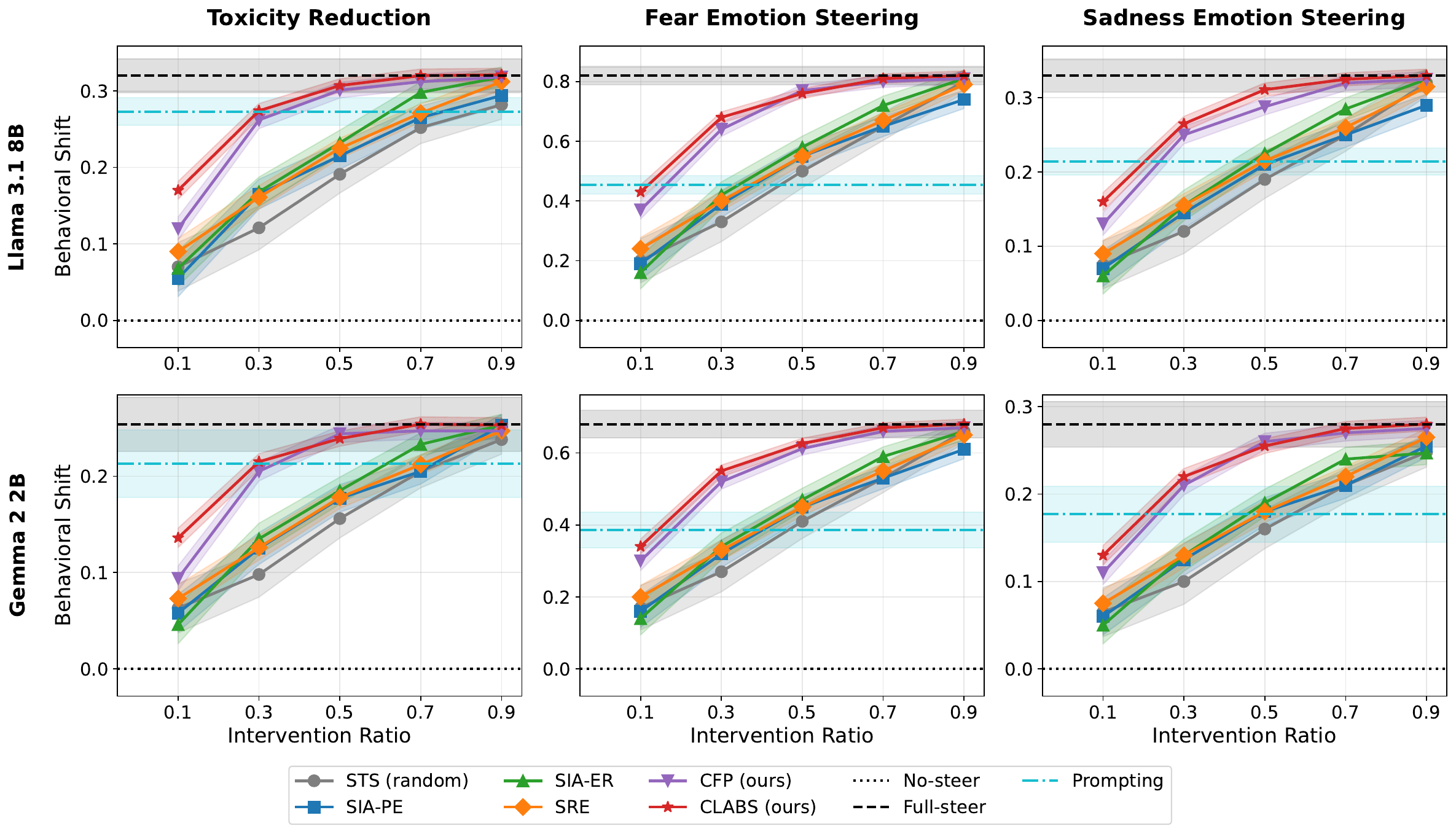}
\caption{\textbf{Behavioral shift versus intervention ratio $p$} on two models
and three targets. \textbf{CFP} and \textbf{CLABS} achieve the largest shift
among the budgeted gates, with the widest margin at low $p$.}
    \label{fig:clabs_behavior_1}
\end{figure*}

\subsection{Baselines}
\label{sec:baselines}
All methods share the steering intervention and differ only in the binary
gate $m_t\in\{0,1\}$ that decides when the shared direction is injected. We
compare against three unbudgeted reference points that bound the achievable
range and four budgeted gates that, like ours, allocate a fixed budget $p$.

\begin{enumerate}
\item \textbf{No-steer} sets $m_t\equiv 0$ and serves as the uncontrolled
reference against which behavioral shift is measured.
\item \textbf{Full-steer} sets $m_t\equiv 1$, the dense fixed-magnitude limit
of our framework.
\item \textbf{Prompting} prepends a task-specific safety or emotion
instruction, refined by iterative prompt engineering, to the input
(Appendix~\ref{appx_examples}).
\end{enumerate}

Adaptive magnitude controllers such as PID Steering \cite{nguyen2026pid}
operate on a complementary axis. They modulate \emph{how much} to steer at
every position rather than \emph{when} to steer, and therefore admit no budget
$p<1$. A gate and a magnitude controller compose naturally, so we do not treat
them as competitors under matched allocation.

Two budgeted gates are derived from Sparse Inference-time Alignment (SIA)
\cite{hu2026sia}. SIA is governed by a reward threshold rather than a token
budget and relies on a learned value model and multi-sample search, so it
cannot be run at a matched budget as published. We therefore extract its
selection signal, the predictive entropy of the next-token distribution, and
turn it into a single-pass gate calibrated to the same $p$. This isolates SIA's
selection hypothesis, that intervention should target tokens where the model is
uncertain, from its scorer and search, and is not a reproduction of the full
method.

\begin{enumerate}
\setcounter{enumi}{3}
\item \textbf{SIA-PE} \cite{hu2026sia} fires on tokens of high predictive
entropy.
\item \textbf{SIA-ER} \cite{hu2026sia} fires on a relative rise in entropy
from the previous token.
\item \textbf{SAE Reconstruction Error (SRE)}, which we introduce as a
mechanistic counterpoint to the uncertainty gates, steers the tokens that the
dictionary represents faithfully,
$m_t=\Ind[\,\norm{h_t-\sae(h_t)}/(\norm{h_t}+\epsilon)\le\tau\,]$.
\item \textbf{STS (random)} \cite{eshraghi2026sts} steers each token
independently with $m_t\sim\mathrm{Bernoulli}(p)$. Its selection is unrelated
to any token property, so it isolates the effect of the budget alone and serves
as the allocation-agnostic control.
\end{enumerate}

Because these gates respond to different signals, a shared threshold would
yield different budgets. We therefore calibrate every gate to the same target
$p$ on a held-out split by tuning its single control parameter under active
steering (Appendix~\ref{app:implementation}). Matching the budget at calibration
time does not by itself guarantee matched spending at test time, so we also
measure the realized intervention rate on the evaluation prompts. Realized
rates stay close to the target for every method, and CFP and CLABS do not spend
more than the gates they outperform (Appendix~\ref{app:results}).

\subsection{Evaluation and Metrics}
\label{sec:metrics}
We assess each method along two axes, behavioral efficacy and the quality
cost at which it is obtained. Efficacy is measured by the \emph{behavioral
shift}, the change in an external classifier's score on the continuation
relative to No-steer. The prompt is excluded, and the shift is signed by task
polarity so that a larger value always means a larger move toward the target.
Toxicity is scored with a RoBERTa classifier and emotion with a DistilRoBERTa
classifier fine-tuned on the Ekman categories \cite{hartmann2022emotionenglish}. Further details are reported in Appendix~\ref{app:implementation}.

Utility is measured by the \textsc{AlpacaEval} win rate \cite{alpaca_eval}
against No-steer under two independent judges, \textsc{gpt-5-mini} and
\textsc{Claude Sonnet 4.6} \citep{openai2025gpt5mini,anthropic2026sonnet46},
with position randomization. A value above $50$ means the steered response is
preferred. We complement the
win rate with perplexity, rep-3, and distinct-2, which track fluency,
degeneration, and diversity of the same responses.

\begin{table*}[t]
\centering
\caption{\textbf{Instruction following and generation quality on AlpacaEval under
\emph{toxicity-reduction} steering.} Win rate is the fraction of instructions where the steered
response beats No-steer. PPL, rep-3, and distinct-2 track fluency and
diversity. CFP and CLABS retain utility as $p$ grows, while the other gates and
Full-steer lose quality.}
\label{tab:main}
\begin{tabular}{llccccc}
\toprule
Method & $p$ & \multicolumn{2}{c}{Win Rate $\uparrow$} & PPL $\downarrow$ & rep-3 $\downarrow$ & distinct-2 $\uparrow$ \\
\cmidrule(lr){3-4}
       &     & GPT-5-mini & Sonnet 4.6 & & & \\
\midrule
No-Steer   & 0   & --  & -- & $23.0_{\pm 2.4}$ & $0.01_{\pm 0.004}$ & $\mathbf{0.91_{\pm 0.03}}$ \\
\midrule
Prompting  & --   & $42.4_{\pm 2.6}$ & $46.3_{\pm 2.6}$ & $23.5_{\pm 2.5}$ & $0.01_{\pm 0.005}$ & $0.90_{\pm 0.03}$ \\
\midrule
Full-Steer & 1.0  & $43.2_{\pm 2.9}$ & $44.9_{\pm 2.8}$ & $41.2_{\pm 6.8}$ & $0.09_{\pm 0.03}$  & $0.68_{\pm 0.07}$ \\
\midrule
\multirow{3}{*}{SIA-ER}
           & 0.2 & $52.1_{\pm 2.7}$ & ${56.7_{\pm 2.6}}$ & $25.1_{\pm 2.7}$ & $0.02_{\pm 0.007}$ & $0.88_{\pm 0.04}$ \\
           & 0.5 & $50.8_{\pm 2.8}$ & $55.0_{\pm 2.7}$ & $29.4_{\pm 3.6}$ & $0.04_{\pm 0.01}$  & $0.82_{\pm 0.05}$ \\
           & 0.8 & $44.3_{\pm 3.1}$ & $49.5_{\pm 2.9}$ & $38.7_{\pm 6.1}$ & $0.08_{\pm 0.03}$  & $0.72_{\pm 0.07}$ \\
\midrule
\multirow{3}{*}{SIA-PE}
           & 0.2 & ${58.6_{\pm 2.5}}$ & ${56.7_{\pm 2.5}}$ & $24.8_{\pm 2.6}$ & $0.02_{\pm 0.006}$ & $0.89_{\pm 0.04}$ \\
           & 0.5 & $54.0_{\pm 2.7}$ & $54.0_{\pm 2.7}$ & $28.1_{\pm 3.3}$ & $0.03_{\pm 0.01}$  & $0.84_{\pm 0.05}$ \\
           & 0.8 & $47.1_{\pm 3.0}$ & $49.2_{\pm 2.9}$ & $36.2_{\pm 5.7}$ & $0.07_{\pm 0.02}$  & $0.74_{\pm 0.06}$ \\
\midrule
\multirow{3}{*}{SRE}
           & 0.2 & $51.2_{\pm 2.6}$ & $53.5_{\pm 2.6}$ & $24.3_{\pm 2.6}$ & $0.02_{\pm 0.006}$ & $0.89_{\pm 0.04}$ \\
           & 0.5 & $53.4_{\pm 2.6}$ & $54.7_{\pm 2.6}$ & $27.0_{\pm 3.1}$ & $0.03_{\pm 0.01}$  & $0.85_{\pm 0.05}$ \\
           & 0.8 & $49.0_{\pm 2.9}$ & $49.6_{\pm 2.8}$ & $33.5_{\pm 5.0}$ & $0.06_{\pm 0.02}$  & $0.77_{\pm 0.06}$ \\
\midrule
\multirow{3}{*}{STS (random)}
           & 0.2 & $51.0_{\pm 2.7}$ & $53.2_{\pm 2.7}$ & $25.6_{\pm 2.8}$ & $0.02_{\pm 0.007}$ & $0.88_{\pm 0.04}$ \\
           & 0.5 & $50.1_{\pm 2.8}$ & $52.3_{\pm 2.8}$ & $30.2_{\pm 3.8}$ & $0.04_{\pm 0.01}$  & $0.81_{\pm 0.05}$ \\
           & 0.8 & $43.5_{\pm 3.2}$ & $48.5_{\pm 3.0}$ & $39.9_{\pm 6.4}$ & $0.09_{\pm 0.03}$  & $0.70_{\pm 0.07}$ \\
\midrule
\multirow{3}{*}{CFP (ours)}
           & 0.2 & $53.0_{\pm 2.6}$ & $56.4_{\pm 2.5}$ & $23.1_{\pm 2.4}$ & $0.01_{\pm 0.004}$ & $0.90_{\pm 0.03}$ \\
           & 0.5 & $57.1_{\pm 2.5}$ & $56.4_{\pm 2.5}$ & $24.5_{\pm 2.6}$ & $0.02_{\pm 0.006}$ & $0.89_{\pm 0.04}$ \\
           & 0.8 & $52.2_{\pm 2.7}$ & $53.8_{\pm 2.6}$ & $27.8_{\pm 3.2}$ & $0.03_{\pm 0.01}$  & $0.86_{\pm 0.04}$ \\
\midrule
\multirow{3}{*}{CLABS (ours)}
           & 0.2 & $52.5_{\pm 2.6}$ & $56.4_{\pm 2.5}$ & ${22.8_{\pm 2.3}}$ & ${0.01_{\pm 0.003}}$ & ${0.91_{\pm 0.03}}$ \\
           & 0.5 & $56.2_{\pm 2.5}$ & $55.6_{\pm 2.5}$ & $23.9_{\pm 2.5}$ & $0.01_{\pm 0.004}$ & $0.90_{\pm 0.03}$ \\
           & 0.8 & $51.0_{\pm 2.8}$ & $52.6_{\pm 2.6}$ & $26.5_{\pm 3.0}$ & $0.02_{\pm 0.006}$ & $0.88_{\pm 0.04}$ \\
\bottomrule
\end{tabular}
\end{table*}

\subsection{Results}
\label{sec:results}

\paragraph{Behavioral efficacy.}
Figure~\ref{fig:clabs_behavior_1} reports the behavioral shift on all six
model--target panels, with shaded bands denoting the standard deviation. At matched budgets CFP and CLABS achieve the largest shift among the
budgeted gates on every panel. Their margin is widest at the smallest budgets,
where a scarce allowance must be spent well, and there they reach two to three times the shift of the uncertainty gates and the random control while steering the
same number of tokens. \emph{Which} tokens are steered, and not only how many,
therefore governs efficacy. As the budget grows the gates converge, as expected
when nearly every token may be steered. Already at a moderate budget CLABS
recovers $89$--$92\%$ of the Full-steer shift on the emotion targets, so
steering about half of the tokens attains most of the effect of steering all of
them. Both of our gates also surpass Prompting, whose shift stays fixed, once
the budget is no longer minimal.

The lead of CLABS at small budgets follows from its design. The integral term
accumulates the running misalignment of the generation, raising the steering
probability while a span drifts off target and relaxing it once alignment is
restored, so the budget concentrates on the spans that actually lag. A
memoryless gate cannot allocate this way and may spend budget on a token that
dips briefly inside a span already on target, and the uncertainty gates fire on
entropy that is only loosely tied to the target behavior, which is why SIA-PE
and SIA-ER fall furthest behind when the budget is tight. The pattern holds on
both a 2B and an 8B model from different families, and the ranking is unchanged
when methods are compared at their realized rather than nominal budgets
(Appendix~\ref{app:results}).

\paragraph{Instruction following.}
Behavioral control is useful only if the model stays helpful. We steer
\textsc{Gemma\,2\,2B} with the toxicity-reduction direction on AlpacaEval
instructions and compare each steered response with its No-steer counterpart
(Table~\ref{tab:main}). CFP and CLABS are the only gates preferred to No-steer
at every budget under both judges. Their fluency, repetition, and diversity
degrades the least, whereas the
other gates lose quality steadily as $p$ grows and Full-steer degrades
markedly. The two judges differ in spread but agree on this ranking.

A single high win-rate cell should be read together with efficacy. SIA-PE
attains the highest \textsc{gpt-5-mini} win rate at the smallest budget, yet
its behavioral shift in that regime is among the weakest
(Figure~\ref{fig:clabs_behavior_1}), its quality erodes quickly as $p$ grows,
and Sonnet scores it slightly above our gates. This reflects a light touch rather
than effective control. The relevant criterion is to sustain efficacy and
utility jointly across budgets and judges, and among the gating rules we
compare only CFP and CLABS meet it. Because every gate shares the same
direction, strength, and corrections, this difference is attributable to
allocation. Sparse, well-placed steering leaves most of the trajectory
unperturbed, whereas STS spends its budget irrespective of need and Full-steer
gives up temporal sparsity altogether. 

Further results, including realized budgets, ablations, and additional instruction-following settings, are given in Appendix~\ref{app:results}, and limitations are discussed in Appendix~\ref{appx:limitations}.

\subsection{Discussion}\label{sec:discussion}

\paragraph{Why allocation and history matter.} If every steering step shifted alignment equally, any allocation of a fixed budget would reach the same mean behavior. Allocation matters because the effect of steering varies across tokens, as the consistent gap between the informed gates and random gating at matched realized budgets shows. CFP and CLABS read the same alignment signal, so the gap between them reflects the value of memory. The leaky integral weights persistent errors more than fluctuating ones, so CLABS spends more of its budget on sustained drift than on momentary dips, which a memoryless gate treats alike. This should matter most when the budget is scarce, where CLABS leads in Figure~\ref{fig:clabs_behavior_1}.

\paragraph{Practical guidance.} Neither gate needs an SAE pass, an extra forward pass, or multi-sample decoding. CFP adds one inner product per token and CLABS a few scalar operations on top of it, which is on the order of $10^{-6}$ of the cost of a forward pass and is therefore negligible. CLABS is preferable under tight budgets, and CFP is a simpler, competitive choice at larger budgets, where the informed gates converge. 
\section{Conclusion}
We framed activation steering as budgeted allocation, in which a per-token gate
decides \emph{when} to spend a fixed intervention budget, so that allocation is the
sole variable. We introduced two gates. CFP is informed but memoryless and steers
the least-aligned tokens by an interpretable feature signal. CLABS is trajectory
aware and spends budget where a generation lags. On a linear steering-response
model, its PI controller is provably stable within an explicit gain region, and
its integral term reduces the steady-state offset. Across two models and three
targets, both gates achieve the largest behavioral shift among budgeted gates, CLABS most clearly at the tight budgets where allocation matters most.
They also best preserve fluency compared to the budgeted gates and full-steering.

\bibliography{aaai2027}

  \clearpage
  \appendix
  \begin{center}
\fbox{\parbox{0.8\columnwidth}{\small\textbf{Content Warning.}
This paper studies inference-time control of language model behavior, including the reduction of toxic generation. The supplementary material contains examples of toxic, offensive, and emotionally distressing model outputs that are shown to illustrate the method. Reader discretion is advised.}}
\end{center}
  \input{appx/thm_proof}
  \input{appx/implementation_details}
  \input{appx/additional_res}
 \newpage
 \clearpage
  \input{appx/qualitative_examples}
  \input{appx/limitations}

\end{document}

%% file: appx/thm_proof.tex
\section{Theoretical Results}\label{appx_thm_proof}

This appendix gives the derivations and proofs behind
Section~\ref{sec:theory}. We first derive the closed-loop recursion,
including the noise contributed by the Bernoulli gate
(Section~\ref{app:noise}). We then prove the stability condition of
Theorem~\ref{thm:stability} (Section~\ref{app:jury}), its three claims
(Section~\ref{app:claims}), Corollary~\ref{cor:integral}, and the offset bound
of Remark~\ref{rem:gains} (Section~\ref{app:cor}). Finally, we present the
error weighting that underlies the role of history in CLABS
(Section~\ref{app:drift}).

All guarantees concern the linear steering-response model of
Assumption~\ref{assumption:plant} in the interior regime, where the clip in
Eq.~\eqref{eq:pi} is inactive. They certify that the gated loop is stable on
this model. They are not worst-case guarantees for the full nonlinear decoding
process, and the efficacy of CLABS at matched budgets is assessed empirically.

\paragraph{Notation.}
We recall the quantities of Section~\ref{sec:method}. The standardized
alignment is $\hat s_t$, the setpoint is $s^\star$, the tracking error is
$e_t=s^\star-\hat s_t$, and its leaky integral is
$\bar e_t=(1-\beta)\bar e_{t-1}+\beta e_t$ with $\bar e_0=0$. In the interior
regime the steering probability is $p_t=p_0+\kappa_pe_t+\kappa_i\bar e_t$, and
the gate is $m_t\sim\mathrm{Bernoulli}(p_t)$. The history
$\mathcal F_{t-1}$ contains everything observed before $p_t$ is computed, so
$\hat s_t$, $e_t$, $\bar e_t$, and $p_t$ are all known given
$\mathcal F_{t-1}$. For a vector $z$, $[z]_1$ is its first entry. For a matrix
$M$, $[M]_{11}$ is its top-left entry, $\|M\|$ its operator norm, and
$\rho(M)$ its spectral radius. For symmetric matrices, $M\preceq M'$ means
that $M'-M$ is positive semidefinite.

\subsection{Gate Noise and the Closed-Loop Recursion}
\label{app:noise}

\paragraph{Gate noise.}
The model of Eq.~\eqref{eq:plant} is driven by the realized gate $m_t$. We
split it into its probability and a noise term, $m_t=p_t+\xi_t$. Because $p_t$
is fixed before the coin is flipped,
\begin{equation}
\E[\xi_t\mid\mathcal F_{t-1}]=0,
\label{eq:gate-mean}
\end{equation}
\begin{equation}
\Var(\xi_t\mid\mathcal F_{t-1})=p_t(1-p_t)\le\tfrac14 .
\label{eq:gate-noise}
\end{equation}
The gate therefore acts as a zero-mean disturbance and introduces no bias.

\paragraph{Effective noise.}
Writing $m_t=p_t+\xi_t$ in Eq.~\eqref{eq:plant} gives
\begin{equation}
\hat s_{t+1}=\phi\,\hat s_t+(1-\phi)\beta_0+g\,p_t+\tilde\eta_t,
\label{eq:plant-p}
\end{equation}
with effective noise $\tilde\eta_t=\eta_t+g\,\xi_t$. Both parts have zero mean
given $\mathcal F_{t-1}$, so $\tilde\eta_t$ does too. If $\eta_t$ and $\xi_t$
are conditionally uncorrelated, their variances add, and
\begin{equation}
\Var(\tilde\eta_t\mid\mathcal F_{t-1})
\le\bar\sigma^2:=\sigma_\eta^2+\tfrac{g^2}{4}.
\label{eq:sigma-bar}
\end{equation}
Without this assumption, the Cauchy--Schwarz inequality gives the bound
$(\sigma_\eta+g/2)^2$ instead, and every result below holds with that
constant in place of $\bar\sigma^2$.

\paragraph{Error recursion.}
Substituting $\hat s_t=s^\star-e_t$ and the PI law into
Eq.~\eqref{eq:plant-p} and collecting terms gives
\begin{equation}
e_{t+1}=a\,e_t-g\kappa_i\,\bar e_t+N-\tilde\eta_t,
\label{eq:e-rec}
\end{equation}
where
\begin{equation}
a:=\phi-g\kappa_p,\qquad
N:=(1-\phi)(s^\star-\beta_0)-g\,p_0 .
\label{eq:a-c}
\end{equation}

\paragraph{Integral recursion.}
Inserting Eq.~\eqref{eq:e-rec} into
$\bar e_{t+1}=(1-\beta)\bar e_t+\beta e_{t+1}$ gives
\begin{equation}
\bar e_{t+1}=\beta a\,e_t
+\big[(1-\beta)-\beta g\kappa_i\big]\bar e_t
+\beta N-\beta\tilde\eta_t .
\label{eq:ebar-rec}
\end{equation}

\paragraph{Matrix form.}
Stacking Eqs.~\eqref{eq:e-rec} and~\eqref{eq:ebar-rec} for
$x_t=(e_t,\bar e_t)^\top$ yields Eq.~\eqref{eq:state}, with
\begin{equation}
b=N\,r,\qquad w_t=-r\,\tilde\eta_t,\qquad r:=(1,\beta)^\top .
\label{eq:b-w}
\end{equation}
Both rows are driven by the same scalar $\tilde\eta_t$, so $w_t$ has rank-one
covariance. By Eq.~\eqref{eq:sigma-bar}, it satisfies
\begin{equation}
\E[w_t\mid\mathcal F_{t-1}]=0,\qquad
\Cov(w_t\mid\mathcal F_{t-1})\preceq\bar\Sigma_w,
\label{eq:w-bound}
\end{equation}
where $\bar\Sigma_w:=\bar\sigma^2\,rr^\top$. Finally, reading off the entries
of $A$,
\begin{equation}
\det A=(1-\beta)\,a,
\label{eq:det}
\end{equation}
\begin{equation}
\operatorname{tr}(A)=a+(1-\beta)-\beta g\kappa_i .
\label{eq:det-tr}
\end{equation}

\subsection{Stability Condition via the Jury Test}
\label{app:jury}

The eigenvalues of $A$ are the roots of the characteristic polynomial
\begin{equation}
P(\lambda)=\lambda^2-\operatorname{tr}(A)\,\lambda+\det A .
\end{equation}
For a real quadratic of this form, the Jury (Schur--Cohn)
criterion~\citep{astrom1997computer} states that both roots lie strictly
inside the unit circle, that is $\rho(A)<1$, if and only if
\begin{equation}
P(1)>0,\qquad P(-1)>0,\qquad |\det A|<1 .
\label{eq:jury-triple}
\end{equation}
We check the three conditions in turn under Eq.~\eqref{eq:jury-cond}.

\emph{Determinant.} Since $0\le a<1$ and $\beta\in(0,1]$,
Eq.~\eqref{eq:det} gives $\det A\in[0,1)$.

\emph{Value at $\lambda=1$.} Using Eqs.~\eqref{eq:det}
and~\eqref{eq:det-tr},
\begin{equation}
P(1)=\beta\,(1-a+g\kappa_i).
\end{equation}
This is positive because $\beta>0$, $a<1$, and $g\kappa_i\ge0$.

\emph{Value at $\lambda=-1$.} Similarly,
\begin{equation}
P(-1)=(2-\beta)(1+a)-\beta g\kappa_i ,
\end{equation}
which is positive exactly when
$g\kappa_i<(2-\beta)(1+a)/\beta$, the second inequality of
Eq.~\eqref{eq:jury-cond}.

All three conditions hold, so $\rho(A)<1$. The condition is sufficient but
not necessary. The Jury test also admits some $a<0$, but we restrict to
$a\ge0$ because it gives the simple box of Eq.~\eqref{eq:jury-cond} and keeps
the proportional response from flipping the sign of the error.
\hfill$\square$

\subsection{Proof of the Three Claims of Theorem~\ref{thm:stability}}
\label{app:claims}

With $\rho(A)<1$ established, the three claims follow from classical
arguments for stable linear systems driven by martingale-difference noise. We
give them in full so that the constants are explicit. Because
$\rho(A)<1$, for every $\bar\rho\in(\rho(A),1)$ there is a constant $C>0$ with
\begin{equation}
\|A^t\|\le C\,\bar\rho^{\,t}\qquad\text{for all }t\ge0 .
\label{eq:Apow}
\end{equation}

\paragraph{Claim 1 (mean).}
Taking expectations in Eq.~\eqref{eq:state} and using $\E[w_t]=0$ gives
$\E[x_{t+1}]=A\,\E[x_t]+b$. Since $\rho(A)<1$, the matrix $I-A$ is
invertible, and $x_{\mathrm{ss}}=(I-A)^{-1}b$ is the unique solution of
$x=Ax+b$. The deviation $d_t=\E[x_t]-x_{\mathrm{ss}}$ then obeys
$d_{t+1}=A\,d_t$, so $d_t=A^td_0$. By Eq.~\eqref{eq:Apow},
\begin{equation}
\|\E[x_t]-x_{\mathrm{ss}}\|\le C\,\bar\rho^{\,t}\,\|d_0\|,
\end{equation}
which is geometric convergence with rate arbitrarily close to $\rho(A)$.

\paragraph{Claim 2 (variance).}
Let $\tilde x_t=x_t-\E[x_t]$ and $\Sigma_t=\Cov(x_t)$. Subtracting the mean
recursion from Eq.~\eqref{eq:state} gives $\tilde x_{t+1}=A\tilde x_t+w_t$.
Since $\tilde x_t$ is known given $\mathcal F_{t-1}$ and $w_t$ has zero
conditional mean,
\begin{equation}
\E[\tilde x_t w_t^\top]
=\E\big[\tilde x_t\,\E[w_t^\top\mid\mathcal F_{t-1}]\big]=0 .
\end{equation}
The cross terms therefore vanish, and
\begin{equation}
\Sigma_{t+1}=A\Sigma_tA^\top+Q_t,\qquad Q_t:=\E[w_tw_t^\top].
\label{eq:cov-rec}
\end{equation}
By the tower property and Eq.~\eqref{eq:w-bound},
$Q_t=\E[\Cov(w_t\mid\mathcal F_{t-1})]\preceq\bar\Sigma_w$. Define the
comparison sequence $\bar\Sigma_{t+1}=A\bar\Sigma_tA^\top+\bar\Sigma_w$ with
$\bar\Sigma_0=\Sigma_0$. The map $X\mapsto AXA^\top$ preserves $\preceq$, so
induction gives $\Sigma_t\preceq\bar\Sigma_t$ for all $t$. Unrolling the
comparison sequence,
\begin{equation}
\bar\Sigma_t=A^t\Sigma_0(A^\top)^t
+\bar\sigma^2\sum_{k=0}^{t-1}A^krr^\top(A^\top)^k .
\label{eq:sigma-unroll}
\end{equation}
The first term vanishes as $t\to\infty$ by Eq.~\eqref{eq:Apow}. The top-left
entry of the second term is $\bar\sigma^2\sum_{k<t}([A^kr]_1)^2$, which
increases to $c(A)\,\bar\sigma^2$, with $c(A)$ as defined in
Theorem~\ref{thm:stability}. Since $\Var(e_t)=[\Sigma_t]_{11}
\le[\bar\Sigma_t]_{11}$,
\begin{equation}
\limsup_{t\to\infty}\Var(e_t)\le c(A)\,\bar\sigma^2 .
\end{equation}
The series defining $c(A)$ converges by Eq.~\eqref{eq:Apow}, and
$c(A)\ge1$ because its $k=0$ term is $([r]_1)^2=1$. Equivalently,
$c(A)\bar\sigma^2=[\bar\Sigma_\infty]_{11}$, where $\bar\Sigma_\infty$ is the
unique solution of the discrete Lyapunov equation
$\bar\Sigma_\infty=A\bar\Sigma_\infty A^\top+\bar\Sigma_w$. The multiplier is
finite inside the stability box but grows as $\rho(A)\to1$, so the bound is
not uniform in the gains.

\paragraph{Claim 3 (time-averaged tracking).}
Since $e_{\mathrm{ss}}$ is deterministic, the mean squared error splits into a
squared bias and a variance,
\begin{equation}
\E\big[(e_t-e_{\mathrm{ss}})^2\big]
=\big(\E[e_t]-e_{\mathrm{ss}}\big)^2+\Var(e_t).
\end{equation}
By Claim~1 the squared bias is at most $C^2\|d_0\|^2\bar\rho^{\,2t}$. By
Eq.~\eqref{eq:sigma-unroll} the variance is at most
$C^2\|\Sigma_0\|\bar\rho^{\,2t}+c(A)\,\bar\sigma^2$. Summing the geometric
terms over $t=1,\dots,T$ and dividing by $T$,
\begin{align}
\frac1T\sum_{t=1}^T\E\big[(e_t-e_{\mathrm{ss}})^2\big]
&\le\frac{K}{T}+c(A)\,\bar\sigma^2,
\label{eq:rate-explicit}\\
K&:=\frac{C^2\big(\|d_0\|^2+\|\Sigma_0\|\big)\bar\rho^{\,2}}
{1-\bar\rho^{\,2}} .
\notag
\end{align}
The first term is the transient, which decays as $O(1/T)$, and the second is
the noise floor, which is $O(\bar\sigma^2)$. This is
Eq.~\eqref{eq:rate}. \hfill$\square$

\subsection{Proof of Corollary~\ref{cor:integral} and Remark~\ref{rem:gains}}
\label{app:cor}

The noise has zero mean, so the deterministic steady state is the fixed point
$x_{\mathrm{ss}}=(e_{\mathrm{ss}},\bar e_{\mathrm{ss}})^\top$ of $x=Ax+b$,
which exists and is unique because $\rho(A)<1$. The integral row reads
$\bar e_{\mathrm{ss}}=(1-\beta)\bar e_{\mathrm{ss}}+\beta e_{\mathrm{ss}}$,
so $\bar e_{\mathrm{ss}}=e_{\mathrm{ss}}$ since $\beta>0$. In steady state the
leaky integral equals the error. Substituting this into the error row,
Eq.~\eqref{eq:e-rec} without noise, gives
$e_{\mathrm{ss}}=a\,e_{\mathrm{ss}}-g\kappa_i e_{\mathrm{ss}}+N$, and hence
\begin{equation}
e_{\mathrm{ss}}=\frac{N}{D},\qquad
D:=1-a+g\kappa_i=1-\phi+g(\kappa_p+\kappa_i),
\label{eq:ess-app}
\end{equation}
with $N$ as in Eq.~\eqref{eq:a-c}. This is Eq.~\eqref{eq:ess}. Under
Eq.~\eqref{eq:jury-cond}, $1-a>0$ and $g\kappa_i\ge0$, so $D>0$. With $p_0$
fixed, $N$ does not depend on $\kappa_i$, while $D$ strictly increases with
$\kappa_i$ because $g>0$. Therefore $|e_{\mathrm{ss}}|=|N|/D$ is
non-increasing in $\kappa_i$, and strictly decreasing when $N\neq0$.
\hfill$\square$

\paragraph{Offset bound of Remark~\ref{rem:gains}.}
With $\kappa_p=\kappa_i=0$, Eq.~\eqref{eq:ess-app} gives
$e_{\mathrm{ss}}=N/(1-\phi)$, so $N$ is $(1-\phi)$ times the gap between the
setpoint and the alignment that an open-loop gate with bias $p_0$ reaches.
For fixed $p_0$ and $\kappa_p$, the second inequality of
Eq.~\eqref{eq:jury-cond} gives $g\kappa_i<(2-\beta)(1+a)/\beta$, and hence,
for $N\neq0$,
\begin{equation}
|e_{\mathrm{ss}}|=\frac{|N|}{1-a+g\kappa_i}
>\frac{|N|}{1-a+(2-\beta)(1+a)/\beta}>0 .
\label{eq:offset-bound-app}
\end{equation}
This is Eq.~\eqref{eq:offset-bound}. Integral action within the box
therefore reduces the offset but does not remove it. \hfill$\square$

The corollary and the bound compare controllers at a \emph{fixed} bias $p_0$.
In our experiments $p_0$ is recalibrated for every gain setting to match the
budget, which changes $N$. These results therefore describe the controller at
a fixed operating point and do not by themselves predict the ranking of gates
at a matched budget, which we assess empirically in
Section~\ref{sec:experiments}.

\subsection{Persistent Versus Alternating Errors}
\label{app:drift}

This result formalizes the claim of Section~\ref{sec:clabs} that a sustained
drift draws more budget than an isolated dip. Unrolling the leaky integral
shows that it is a weighted average of past errors,
\begin{equation}
\bar e_t=\beta\sum_{k=0}^{t-1}(1-\beta)^k e_{t-k}.
\end{equation}
We compare how the steering probability responds to two error patterns of
the same magnitude $|e|$.

\emph{Persistent error.} If $e_t\equiv e$, then $\bar e_t=(1-(1-\beta)^t)\,e
\to e$, and the response $\kappa_pe_t+\kappa_i\bar e_t$ approaches
$(\kappa_p+\kappa_i)\,e$.

\emph{Alternating error.} If $e_t=(-1)^te$, the leaky integral approaches
the periodic solution $\bar e_t=(-1)^t\gamma e$. Substituting into
$\bar e_t=(1-\beta)\bar e_{t-1}+\beta e_t$ gives
$\gamma=-(1-\beta)\gamma+\beta$, so
\begin{equation}
\gamma=\frac{\beta}{2-\beta}.
\end{equation}
The response therefore approaches
$(-1)^t\big(\kappa_p+\kappa_i\beta/(2-\beta)\big)e$. Since
$\beta/(2-\beta)<1$ for every $\beta\in(0,1)$, a persistent error receives
more weight than an alternating one whenever $\kappa_i>0$, and the gap widens
as the memory lengthens ($\beta\to0^+$). A memoryless gate, with $\kappa_i=0$,
gives both patterns the same weight.

%% file: appx/implementation_details.tex
\section{Implementation Details}
\label{app:implementation}

This appendix describes the models, steering directions, decoding settings, and per-method hyperparameters. It then covers the budget calibration protocol and the computational overhead. Within each (model, task) pair, all methods share the same steering direction, intervention layer, per-token strength, decoding settings, and post-intervention corrections. The gating policy is therefore the only variable.

\paragraph{Models, Directions, and Decoding}

We evaluate two open base models, \textsc{Gemma\,2\,2B} \cite{team2024gemma2} and
\textsc{Llama\,3.1\,8B} \cite{dubey2024llama}. For \textsc{Gemma\,2\,2B} we use
the Gemma Scope residual-stream SAE \cite{lieberum2024gemmascope}. For
\textsc{Llama\,3.1\,8B} we use the EleutherAI residual-stream
SAE.\footnote{\url{https://huggingface.co/EleutherAI/sae-llama-3.1-8b-32x}}
In both models we read the SAE and inject the steering direction at the
residual stream of the $16$th transformer layer only. This follows the common
practice of steering at a single mid-network layer \cite{labonne2024clamping}.

For every (model, task) pair we extract one steering direction $v$ from a
target corpus $\mathcal D^+$ and a neutral corpus $\mathcal D^-$
(Eq.~\ref{eq:direction}). For elicitation, $\mathcal D^+$ exhibits the emotion
to be induced. For suppression, it exhibits the behavior to be removed (toxic
text), and the negative strength $\alpha<0$ moves activations away from it.
Each $f_i(x)$ in Eq.~\eqref{eq:differential} is the SAE code of feature $i$
averaged over the tokens of $x$. We take the top-$k$ features ($k=10$) ranked
by $\Delta_i$, then sum and normalize their decoder rows. We use $200$
examples per side to build the direction, $100$ held-out examples for
calibration, and $300$ for evaluation. The three splits are disjoint. A
constant $\epsilon=10^{-6}$ is used for numerical stability throughout, for
example in Eq.~\eqref{eq:s-standardized}.

For toxicity reduction, \textsc{RealToxicityPrompts} supplies per-example
toxicity scores. We therefore form high-purity endpoints directly, with target
texts of toxicity $\ge0.9$ and neutral texts of toxicity $\le0.2$.
\textsc{GoEmotions} provides categorical labels without graded intensities.
For emotion steering we therefore score every candidate text with the emotion
classifier. We then sample the neutral, fear, and sadness corpora from texts
whose classifier probability for the relevant category exceeds $0.9$. These
direction corpora are disjoint from all evaluation prompts.

For emotion prompts we truncate each neutral input to $50\%$ of its token
length before generation. This way the continuation reflects the intervention
rather than residual emotional content in the prompt. For toxicity we keep the
full prompt, since truncation would remove the toxic context that makes
detoxification hard. Toxicity is scored by
\texttt{s-nlp/roberta\_toxicity\_classifier}.\footnote{\url{https://huggingface.co/s-nlp/roberta_toxicity_classifier}}
Emotion is scored by \texttt{j-hartmann/emotion-english-distilroberta}
\cite{hartmann2022emotionenglish},\footnote{\url{https://huggingface.co/j-hartmann/emotion-english-distilroberta-base}}
which is fine-tuned on Ekman's six emotions plus neutral. Behavioral shift is
computed on the continuation only. Generation uses \texttt{max\_new\_tokens}$=128$, temperature $0.1$, and a
repetition penalty of $1.1$ for all methods. Three method-agnostic corrections
are applied identically to every steered token.
\begin{itemize}
  \item \emph{Norm preservation} \cite{turner2023activation} rescales the
    steered activation to its pre-intervention norm. The intervention thus
    rotates toward $v$ without inflating magnitude.
  \item \emph{Activation clamping} \cite{labonne2024clamping} bounds the
    scalar projection onto $v$ to $[0,40]$. This prevents over-amplification
    on tokens that are already aligned.
  \item A decoding-stage \emph{repetition penalty} \cite{keskar2019ctrl}
    discounts the logits of previously generated tokens. This counteracts the
    repetition that suppressing high-probability continuations can induce.
\end{itemize}

Steering applies $\tilde h_t=h_t+m_t\alpha v$ at the fixed layer. The strength
$\alpha$ is fixed per (model, task) and is identical across all gates.
Following common practice \cite{labonne2024clamping,eshraghi2026sts}, we
select $\alpha$ as the largest magnitude that preserves fluency on the
calibration split. For \textsc{Gemma\,2\,2B} we use $\alpha=-35, 30, 30$ for
toxicity, fear, and sadness. For \textsc{Llama\,3.1\,8B} we use
$\alpha=-4,3,2.5$. Negative values suppress and positive values elicit. All
experiments were run on NVIDIA L4 and A10 GPUs.

\paragraph{Gate signals and timing.}
CFP, CLABS, and SRE read the residual stream at layer $\ell$ before the
intervention at the current step. SIA-PE and SIA-ER use the entropy of the
next-token distribution from which the current input token was sampled. This
entropy is computed at the end of the previous step. The next-token
distribution of the current step exists only after the full forward pass,
which comes after the intermediate-layer intervention. Every gate is therefore
causal and single-pass. The cost is a one-step lag for the entropy gates,
which any single-pass entropy rule must accept.

\paragraph{Method hyperparameters.}
The threshold gates (SIA-PE, SIA-ER, SRE, and CFP) have no hyperparameter
beyond the calibrated threshold $\tau$. STS (random) fires each token
independently with probability $p$ \cite{eshraghi2026sts}. CLABS uses gains
$(\kappa_p,\kappa_i,\beta)=(1.0,0.2,0.2)$ and setpoint $s^\star=0.75$ in
standardized units. The statistics $\mu$ and $\sigma$ in
Eq.~\eqref{eq:s-standardized} are computed on the calibration split. With
$\beta=0.2$, the leaky integral averages roughly the last five tokens.

\paragraph{Budget calibration.}
Every budgeted method is calibrated to the same target rate $p$, so that all
comparisons are made at equal intervention cost. Each method exposes a single
control parameter that sets its firing rate. This parameter is the threshold
$\tau$ for the threshold gates and the bias $p_0$ for CLABS. We tune only this
parameter by bisection and leave all other settings fixed. STS meets the
budget in expectation by construction and needs no tuning. Calibration runs on
the held-out calibration split and never touches the evaluation data. It is
performed under active steering. An unsteered pass would under-spend the
budget, because each intervention lowers the alignment that later firings of
the alignment-sensitive gates respond to. Evaluation then uses fresh seeds.
The same procedure is repeated for every budget, every ablation variant, and
every cell of the gain sweep. Realized test-time rates on the evaluation
prompts are reported in Appendix~\ref{app:rates}.

\paragraph{Computational overhead.}
The probe $u$ is a fixed vector, so neither gate requires an SAE pass at
inference. CFP computes one inner product per token, which costs $2d$ FLOPs
with $d=2304$ for Gemma and $d=4096$ for Llama. We estimate this at roughly
$5$ to $8$ kFLOPs per token. A forward pass costs roughly $5$ to $16$ GFLOPs
per token, taking about two FLOPs per parameter. The relative cost is thus on
the order of $10^{-6}$. CLABS adds a few scalar operations and one Bernoulli
draw. Neither gate requires an extra forward pass, rollouts, backtracking, or
multi-sample decoding. In contrast, the full SIA method relies on a learned
value model and multi-sample search \cite{hu2026sia}.

\paragraph{AlpacaEval protocol.}
The utility experiment samples $200$ instructions from \textsc{AlpacaEval}
\cite{alpaca_eval,dubois2024alpacafarm}. Open-ended instruction following
benefits from more diverse generation. We therefore decode with temperature
$0.7$, \texttt{max\_new\_tokens}$=256$, and repetition penalty $1.1$, and leave
the steering mechanism and all corrections unchanged. Each steered response is
compared with the No-steer response to the same instruction by pairwise
judging. The A/B position is randomized per pair, and ties are scored as
$0.5$. The two judges, \textsc{gpt-5-mini} \cite{openai2025gpt5mini} and
\textsc{Claude Sonnet 4.6} \cite{anthropic2026sonnet46}, are queried at
temperature $0$. Further results and the
judge prompt are given in Appendices~\ref{appx:alpaca_fear}
and~\ref{appx_examples}.

%% file: appx/additional_res.tex
\section{Additional Results}
\label{app:results}

This appendix collects the evidence behind Section~\ref{sec:results}. We first confirm that all methods are compared at matched realized budgets. Next, we isolate the role of the history term with ablations. Finally, we report further instruction following results across models and targets.

\subsection{Realized Intervention Rates}
\label{app:rates}

\begin{table}[h]
\centering
\small
\caption{
Realized test-time intervention rates.}
\label{tab:realized}
\setlength{\tabcolsep}{2.5pt}
\begin{tabular}{lcccccc}
\toprule
$p$ & STS & SIA-ER & SIA-PE & SRE & CFP & CLABS\\
\midrule
$0.1$ & $.09$--$.12$ & $.09$--$.12$ & $.09$--$.11$ & $.08$--$.11$ & $.09$--$.11$ & $.09$--$.11$\\
$0.3$ & $.28$--$.32$ & $.29$--$.32$ & $.29$--$.31$ & $.29$--$.32$ & $.29$--$.31$ & $.28$--$.31$\\
$0.5$ & $.48$--$.53$ & $.49$--$.52$ & $.48$--$.51$ & $.49$--$.51$ & $.48$--$.51$ & $.48$--$.52$\\
$0.7$ & $.67$--$.73$ & $.68$--$.71$ & $.68$--$.72$ & $.69$--$.71$ & $.68$--$.72$ & $.68$--$.71$\\
$0.9$ & $.86$--$.94$ & $.87$--$.93$ & $.88$--$.91$ & $.88$--$.92$ & $.88$--$.92$ & $.87$--$.91$\\
\bottomrule
\end{tabular}
\end{table}

Every gate is calibrated to the target budget on a held-out split, as described in Appendix~\ref{app:implementation}. Calibration fixes the expected rate on that
split, yet a gate that reacts to the steered state could spend differently on
new prompts. We therefore measure the fraction of steered tokens on the
evaluation prompts for every method, budget, and panel.
Table~\ref{tab:realized} reports the range over the six model--target panels.
Realized rates stay within $0.04$ of the nominal budget for all methods. The
widest ranges belong to STS, whose independent per-token draws carry the most
sampling variability, and they widen as the budget grows.
The ranges of CFP and CLABS lie within those of STS at every budget, so they
do not spend more than the baselines they outperform. Reading the curves of
Figure~\ref{fig:clabs_behavior_1} at the realized instead of the nominal rates
leaves the ranking of methods unchanged on every panel.


\subsection{Ablation on the History Term}
\label{appx:ablation}

\begin{table*}[t]
\centering
\small
\caption{\textbf{CLABS component ablation} on \textsc{Gemma\,2\,2B} toxicity
reduction at a matched budget $p=0.5$ (mean $\pm$ std).}
\label{tab:ablation}
\begin{tabular}{llcccc}
\toprule
$p$ & Variant & Tox.\ red.\ $\uparrow$ & PPL $\downarrow$ & rep-3 $\downarrow$ & distinct-2 $\uparrow$ \\
\midrule
-- & No-Steer & -- & $8.4_{\pm 0.2}$ & $0.02_{\pm 0.01}$ & $0.95_{\pm 0.01}$ \\
\midrule
\multirow{4}{*}{0.5}
 & Open-loop  & $0.12_{\pm 0.02}$ & $9.8_{\pm 0.7}$ & $0.02_{\pm 0.01}$ & $0.94_{\pm 0.01}$ \\
 & I-only     & $0.15_{\pm 0.03}$ & $9.0_{\pm 0.5}$ & $0.03_{\pm 0.01}$ & $0.93_{\pm 0.02}$ \\
 & P-only     & $0.21_{\pm 0.02}$ & $8.9_{\pm 0.4}$ & $0.02_{\pm 0.01}$ & ${0.95_{\pm 0.01}}$ \\
 & CLABS (PI) & ${0.24_{\pm 0.02}}$ & ${8.6_{\pm 0.2}}$ & $0.02_{\pm 0.01}$ & $0.94_{\pm 0.01}$ \\
\bottomrule
\end{tabular}
\end{table*}

\begin{figure*}[t]
    \centering
    \includegraphics[width=\textwidth]{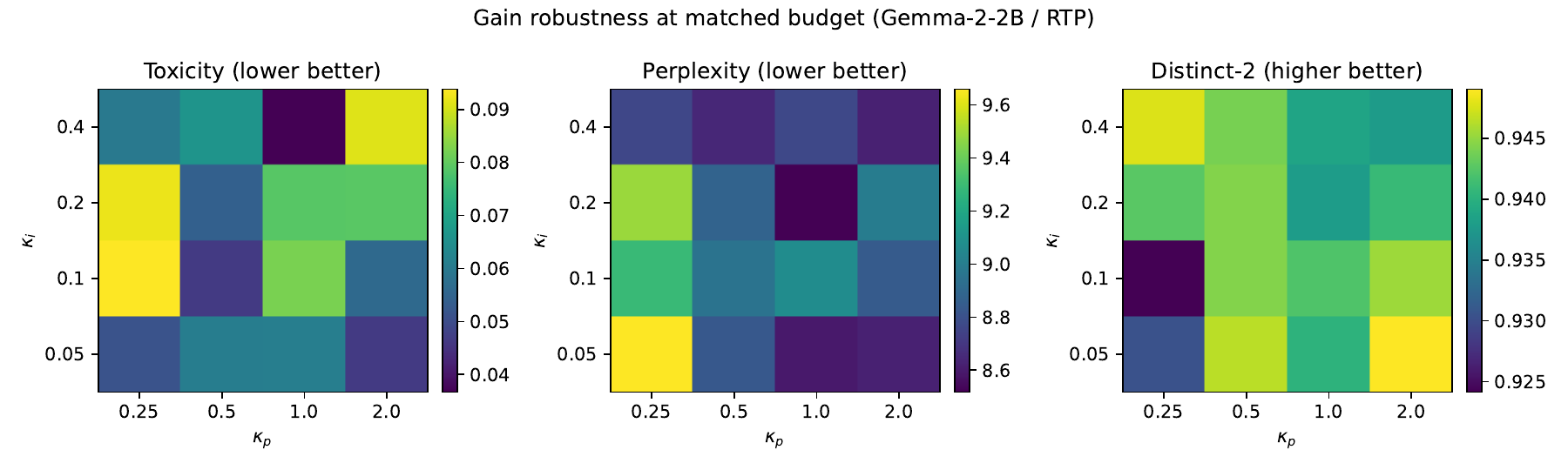}
\caption{\textbf{Gain robustness at matched budget} (\textsc{Gemma\,2\,2B},
toxicity). Toxicity (lower is better), perplexity (lower is better), and
distinct-2 (higher is better) across the $\kappa_p\times\kappa_i$ grid.}
    \label{fig:gain_robustness}
\end{figure*}

\paragraph{Protocol.}
The setup follows Appendix~\ref{app:implementation}, with the same model,
direction, calibration split, and evaluation protocol, averaged over three
seeds. The only change is a shorter generation cap of
$\texttt{max\_new\_tokens}=64$. This keeps the gain sweep tractable and focuses
the measurement on the early tokens, where the gate must decide whether and
when to act. The cap applies to every variant, so it changes absolute
magnitudes but not the comparisons. 

\paragraph{Variants.}
The \emph{open-loop} gate fires with a constant probability $p_t=p_0$ and
ignores feedback. \emph{P-only} sets $\kappa_i=0$ and reacts only to the
current error, which makes it a stochastic counterpart of CFP. \emph{I-only}
sets $\kappa_p=0$ and reacts only to the leaky integral. \emph{CLABS (PI)} uses
both terms at the default gains.

\paragraph{Component ablation.}
Table~\ref{tab:ablation} breaks CLABS into its parts. All variants are
calibrated to the same budget, so they differ only in where along the trajectory
they spend it. The open-loop gate spends blindly and gives the weakest
reduction and the highest perplexity. Each feedback term recovers part of the
gap on its own. The integral-only controller responds to sustained deviations
but not to local ones, and it shows the largest seed variance in toxicity
reduction. The proportional-only controller does better, yet it reads the same
instantaneous signal as CFP and so cannot tell a brief dip from a sustained
drift. The full PI controller gives the strongest reduction and the lowest
perplexity, while repetition and diversity stay within one standard deviation
of the No-Steer level. The proportional term supplies responsiveness and the
leaky integral supplies memory of persistent error, and their combination gives
the best balance of control and fluency in this ablation.
Figure~\ref{fig:gain_robustness} reports toxicity, perplexity, and distinct-2 across the
$\kappa_p\times\kappa_i$ grid, with every cell recalibrated to the same budget.


\begin{table*}[t]
\centering
\caption{\textbf{Instruction following and generation quality on AlpacaEval
under \emph{fear} steering} (\textsc{Gemma\,2\,2B}, mean $\pm$ std). A win rate above $50$ favors steering.}
\label{tab:main_fear}
\begin{tabular}{llccccc}
\toprule
Method & $p$ & \multicolumn{2}{c}{Win Rate $\uparrow$} & PPL $\downarrow$ & rep-3 $\downarrow$ & distinct-2 $\uparrow$ \\
\cmidrule(lr){3-4}
       &     & GPT-5-mini & Sonnet 4.6 & & & \\
\midrule
No-Steer   & 0   & --  & --  & $23.0_{\pm 2.4}$ & $0.01_{\pm 0.004}$ & $0.91_{\pm 0.03}$ \\
\midrule
Prompting  & --   & $50.3_{\pm 2.6}$ & $48.1_{\pm 2.5}$ & $24.1_{\pm 2.5}$ & $0.01_{\pm 0.005}$ & $0.89_{\pm 0.03}$ \\
\midrule
Full-Steer & 1.0  & $38.4_{\pm 3.1}$ & $39.6_{\pm 2.9}$ & $48.6_{\pm 7.4}$ & $0.12_{\pm 0.04}$  & $0.62_{\pm 0.07}$ \\
\midrule
\multirow{3}{*}{SIA-ER}
           & 0.2 & $39.6_{\pm 2.8}$ & $42.3_{\pm 2.7}$ & $27.2_{\pm 3.0}$ & $0.03_{\pm 0.009}$ & $0.85_{\pm 0.05}$ \\
           & 0.5 & $38.2_{\pm 3.0}$ & $40.7_{\pm 2.8}$ & $33.5_{\pm 4.4}$ & $0.06_{\pm 0.02}$  & $0.78_{\pm 0.06}$ \\
           & 0.8 & $37.4_{\pm 3.2}$ & $39.1_{\pm 3.0}$ & $44.1_{\pm 6.8}$ & $0.11_{\pm 0.04}$  & $0.66_{\pm 0.07}$ \\
\midrule
\multirow{3}{*}{SIA-PE}
           & 0.2 & $44.4_{\pm 2.7}$ & $46.5_{\pm 2.6}$ & $26.8_{\pm 2.9}$ & $0.03_{\pm 0.008}$ & $0.86_{\pm 0.04}$ \\
           & 0.5 & $42.6_{\pm 2.9}$ & $45.7_{\pm 2.7}$ & $31.9_{\pm 4.1}$ & $0.05_{\pm 0.02}$  & $0.80_{\pm 0.05}$ \\
           & 0.8 & $40.7_{\pm 3.1}$ & $43.2_{\pm 2.9}$ & $41.5_{\pm 6.3}$ & $0.09_{\pm 0.03}$  & $0.70_{\pm 0.06}$ \\
\midrule
\multirow{3}{*}{SRE}
           & 0.2 & $51.1_{\pm 2.6}$ & $48.7_{\pm 2.5}$ & $26.1_{\pm 2.8}$ & $0.02_{\pm 0.007}$ & $0.87_{\pm 0.04}$ \\
           & 0.5 & $47.3_{\pm 2.7}$ & $47.1_{\pm 2.6}$ & $30.2_{\pm 3.8}$ & $0.04_{\pm 0.01}$  & $0.82_{\pm 0.05}$ \\
           & 0.8 & $38.6_{\pm 3.1}$ & $41.3_{\pm 2.9}$ & $38.4_{\pm 5.9}$ & $0.08_{\pm 0.03}$  & $0.73_{\pm 0.06}$ \\
\midrule
\multirow{3}{*}{STS (random)}
           & 0.2 & $44.7_{\pm 2.7}$ & $47.6_{\pm 2.6}$ & $27.4_{\pm 3.0}$ & $0.03_{\pm 0.009}$ & $0.85_{\pm 0.05}$ \\
           & 0.5 & $40.7_{\pm 2.9}$ & $44.7_{\pm 2.7}$ & $33.1_{\pm 4.3}$ & $0.05_{\pm 0.02}$  & $0.78_{\pm 0.06}$ \\
           & 0.8 & $38.6_{\pm 3.2}$ & $41.2_{\pm 3.0}$ & $44.8_{\pm 6.9}$ & $0.11_{\pm 0.04}$  & $0.66_{\pm 0.07}$ \\
\midrule
\multirow{3}{*}{CFP (ours)}
           & 0.2 & $ {51.6_{\pm 2.5}}$ & ${49.3_{\pm 2.4}}$ & $24.3_{\pm 2.5}$ & $0.02_{\pm 0.006}$ & $0.88_{\pm 0.04}$ \\
           & 0.5 & $46.9_{\pm 2.7}$ & $46.8_{\pm 2.6}$ & $27.1_{\pm 3.1}$ & $0.03_{\pm 0.01}$  & $0.85_{\pm 0.04}$ \\
           & 0.8 & $42.4_{\pm 2.9}$ & $43.9_{\pm 2.7}$ & $32.4_{\pm 4.2}$ & $0.05_{\pm 0.02}$  & $0.80_{\pm 0.05}$ \\
\midrule
\multirow{3}{*}{CLABS (ours)}
           & 0.2 & $48.7_{\pm 2.6}$ & $48.9_{\pm 2.4}$ & ${23.6_{\pm 2.4}}$ & ${0.01_{\pm 0.004}}$ & ${0.89_{\pm 0.03}}$ \\
           & 0.5 & $44.6_{\pm 2.7}$ & $46.2_{\pm 2.6}$ & $25.4_{\pm 2.8}$ & $0.02_{\pm 0.006}$ & $0.87_{\pm 0.04}$ \\
           & 0.8 & $40.8_{\pm 2.9}$ & $43.4_{\pm 2.7}$ & $29.8_{\pm 3.6}$ & $0.04_{\pm 0.01}$  & $0.83_{\pm 0.05}$ \\
\bottomrule
\end{tabular}
\end{table*}

\subsection{Additional Results on Instruction Following}
\label{appx:alpaca_fear}

All utility experiments follow the protocol of Table~\ref{tab:main}. We use 200
AlpacaEval instructions and three seeds, and compare each steered response
pairwise against the No-Steer response with randomized positions
(Appendix~\ref{app:implementation}). A win rate above $50$ means the steered
response is preferred. Each setting requires nearly 20 method--budget configurations, 200 instructions, and three seeds under two judges, or  2k+ judge calls. We therefore evaluate toxicity reduction, the safety-relevant target, on both models, and assign one emotion target to each model, fear to Gemma 2 2B and sadness to Llama 3.1 8B, so that every target is covered on at least one model and every model on both polarities.

Fear is a demanding test. Toxicity reduction largely preserves the intent of
the task, whereas injecting fear can pull a response away from the instruction
it should answer. Win rates are therefore lower for every method
(Table~\ref{tab:main_fear}). Prompting stays near parity. At the smallest
budget, CFP has the highest win rate among the gates under both judges. At the
moderate budget, SRE and CFP lead under both judges, and CLABS is close behind
under Sonnet but trails more clearly under GPT-5-mini. At the largest budget,
CFP retains the highest win rate. The clearest separation is in fluency. At
every budget, CLABS has the lowest perplexity and repetition and the highest
diversity among the gates, with CFP second. Full-Steer and the uncertainty-
and random-gated baselines lose fluency and helpfulness quickly as the budget
grows.

Tables~\ref{tab:llama_tox} and~\ref{tab:llama_sad} extend the study to the
larger model. Under toxicity reduction, CFP and CLABS are the only gates whose
responses are preferred to No-Steer at every budget under both judges. Their
perplexity also grows far more slowly with the budget than that of any other
gate. As on Gemma, the entropy gate scores well under GPT-5-mini at the
smallest budget but loses quality quickly as the budget grows. Sadness, like
fear, is semantically disruptive, and win rates are lower across the board.
Even so, CFP or CLABS has the highest win rate at every budget under both
judges, and CLABS keeps the best fluency, repetition, and diversity among the
gates throughout.

\begin{table*}[t]
\centering
\caption{\textbf{Instruction following and generation quality on AlpacaEval
under \emph{toxicity-reduction} steering} (\textsc{Llama\,3.1\,8B}, mean $\pm$ std). A win rate above $50$ favors
steering.}
\label{tab:llama_tox}
\begin{tabular}{llccccc}
\toprule
Method & $p$ & \multicolumn{2}{c}{Win Rate $\uparrow$} & PPL $\downarrow$ & rep-3 $\downarrow$ & distinct-2 $\uparrow$ \\
\cmidrule(lr){3-4}
       &     & GPT-5-mini & Sonnet 4.6 & & & \\
\midrule
No-Steer   & 0   & --     & --     & $21.4_{\pm 2.2}$ & $0.01_{\pm 0.005}$ & $0.92_{\pm 0.02}$\\
\midrule
Prompting  & --  & $44.3_{\pm 3.1}$ & $47.1_{\pm 2.6}$ & $21.9_{\pm 2.4}$ & $0.01_{\pm 0.005}$ & $0.91_{\pm 0.03}$\\
\midrule
Full-Steer & 1.0 & $43.1_{\pm 2.7}$ & $45.6_{\pm 2.2}$ & $36.9_{\pm 6.1}$ & $0.08_{\pm 0.05}$ & $0.71_{\pm 0.08}$\\
\midrule
\multirow{3}{*}{SIA-ER}
 & 0.2 & $54.6_{\pm 2.9}$ & $55.8_{\pm 2.4}$ & $23.2_{\pm 3.1}$ & $0.02_{\pm 0.008}$ & $0.89_{\pm 0.03}$\\
 & 0.5 & $52.0_{\pm 2.7}$ & $54.1_{\pm 2.3}$ & $27.0_{\pm 3.8}$ & $0.03_{\pm 0.01}$ & $0.84_{\pm 0.05}$\\
 & 0.8 & $46.8_{\pm 2.1}$ & $49.7_{\pm 2.3}$ & $34.6_{\pm 5.2}$ & $0.07_{\pm 0.02}$ & $0.75_{\pm 0.08}$\\
\midrule
\multirow{3}{*}{SIA-PE}
 & 0.2 & $ {57.2_{\pm 2.1}}$ & $55.9_{\pm 1.9}$ & $22.9_{\pm 3.3}$ & $0.02_{\pm 0.008}$ & $0.90_{\pm 0.02}$\\
 & 0.5 & $53.1_{\pm 2.3}$ & $53.6_{\pm 2.5}$ & $26.1_{\pm 4.1}$ & $0.03_{\pm 0.02}$ & $0.85_{\pm 0.05}$\\
 & 0.8 & $47.5_{\pm 2.0}$ & $49.8_{\pm 2.1}$ & $32.7_{\pm 6.6}$ & $0.06_{\pm 0.03}$ & $0.76_{\pm 0.07}$\\
\midrule
\multirow{3}{*}{SRE}
 & 0.2 & $52.4_{\pm 3.1}$ & $53.9_{\pm 2.6}$ & $22.6_{\pm 3.7}$ & $0.02_{\pm 0.009}$ & $0.90_{\pm 0.02}$\\
 & 0.5 & $53.5_{\pm 2.9}$ & $54.2_{\pm 2.6}$ & $25.0_{\pm 4.5}$ & $0.03_{\pm 0.02}$ & $0.86_{\pm 0.04}$\\
 & 0.8 & $49.1_{\pm 2.1}$ & $50.3_{\pm 2.4}$ & $30.6_{\pm 5.4}$ & $0.05_{\pm 0.04}$ & $0.79_{\pm 0.05}$\\
\midrule
\multirow{3}{*}{STS (random)}
 & 0.2 & $52.0_{\pm 2.0}$ & $53.4_{\pm 2.5}$ & $23.6_{\pm 2.1}$ & $0.02_{\pm 0.006}$ & $0.89_{\pm 0.03}$\\
 & 0.5 & $50.6_{\pm 2.9}$ & $52.6_{\pm 2.2}$ & $27.8_{\pm 4.3}$ & $0.04_{\pm 0.02}$ & $0.82_{\pm 0.05}$\\
 & 0.8 & $45.2_{\pm 2.9}$ & $48.9_{\pm 1.9}$ & $35.8_{\pm 6.2}$ & $0.08_{\pm 0.04}$ & $0.73_{\pm 0.08}$\\
\midrule
\multirow{3}{*}{CFP (ours)}
 & 0.2 & $54.8_{\pm 2.5}$ & $56.2_{\pm 2.7}$ & $21.6_{\pm 2.4}$ & $0.01_{\pm 0.004}$ & $0.91_{\pm 0.02}$\\
 & 0.5 & ${57.0_{\pm 2.2}}$ & ${56.0_{\pm 2.6}}$ & $22.8_{\pm 4.0}$ & $0.02_{\pm 0.01}$ & $0.90_{\pm 0.02}$\\
 & 0.8 & ${53.0_{\pm 1.9}}$ & ${54.1_{\pm 2.1}}$ & $25.7_{\pm 4.7}$ & $0.03_{\pm 0.01}$ & $0.87_{\pm 0.03}$\\
\midrule
\multirow{3}{*}{CLABS (ours)}
 & 0.2 & $54.5_{\pm 2.9}$ & ${56.3_{\pm 2.8}}$ & ${21.3_{\pm 2.0}}$ & ${0.01_{\pm 0.004}}$ & ${0.92_{\pm 0.02}}$\\
 & 0.5 & $56.3_{\pm 2.5}$ & $55.4_{\pm 2.6}$ & ${22.3_{\pm 2.8}}$ & ${0.01_{\pm 0.005}}$ & ${0.91_{\pm 0.03}}$\\
 & 0.8 & $52.1_{\pm 2.4}$ & $53.2_{\pm 2.1}$ & ${24.6_{\pm 3.9}}$ & ${0.02_{\pm 0.01}}$ & ${0.88_{\pm 0.05}}$\\
\bottomrule
\end{tabular}
\end{table*}

\begin{table*}[t]
\centering
\caption{\textbf{Instruction following and generation quality on AlpacaEval
under \emph{sadness} steering} (\textsc{Llama\,3.1\,8B}, mean $\pm$ std). A win rate above $50$ favors steering.}
\label{tab:llama_sad}
\begin{tabular}{llccccc}
\toprule
Method & $p$ & \multicolumn{2}{c}{Win Rate $\uparrow$} & PPL $\downarrow$ & rep-3 $\downarrow$ & distinct-2 $\uparrow$ \\
\cmidrule(lr){3-4}
       &     & GPT-5-mini & Sonnet 4.6 & & & \\
\midrule
No-Steer   & 0   & --     & --     & $21.4_{\pm 2.2}$ & $0.01_{\pm 0.005}$ & $0.92_{\pm 0.02}$\\
\midrule
Prompting  & --  & $50.1_{\pm 2.6}$ & $48.4_{\pm 2.4}$ & $22.0_{\pm 2.4}$ & $0.01_{\pm 0.006}$ & $0.91_{\pm 0.02}$\\
\midrule
Full-Steer & 1.0 & $39.0_{\pm 2.9}$ & $41.2_{\pm 2.7}$ & $40.3_{\pm 7.2}$ & $0.10_{\pm 0.05}$ & $0.66_{\pm 0.07}$\\
\midrule
\multirow{3}{*}{SIA-ER}
 & 0.2 & $41.2_{\pm 2.7}$ & $43.9_{\pm 3.1}$ & $24.8_{\pm 2.9}$ & $0.03_{\pm 0.008}$ & $0.86_{\pm 0.04}$\\
 & 0.5 & $39.6_{\pm 2.4}$ & $42.1_{\pm 2.2}$ & $30.4_{\pm 4.0}$ & $0.05_{\pm 0.01}$ & $0.79_{\pm 0.04}$\\
 & 0.8 & $38.3_{\pm 2.5}$ & $40.6_{\pm 2.3}$ & $37.2_{\pm 6.3}$ & $0.09_{\pm 0.03}$ & $0.69_{\pm 0.06}$\\
\midrule
\multirow{3}{*}{SIA-PE}
 & 0.2 & $45.6_{\pm 2.6}$ & $47.4_{\pm 2.4}$ & $24.3_{\pm 2.6}$ & $0.02_{\pm 0.008}$ & $0.87_{\pm 0.03}$\\
 & 0.5 & $43.8_{\pm 2.7}$ & $46.0_{\pm 2.6}$ & $29.1_{\pm 3.7}$ & $0.04_{\pm 0.01}$ & $0.81_{\pm 0.04}$\\
 & 0.8 & $41.5_{\pm 2.7}$ & $43.6_{\pm 2.3}$ & $35.4_{\pm 5.2}$ & $0.08_{\pm 0.02}$ & $0.71_{\pm 0.06}$\\
\midrule
\multirow{3}{*}{SRE}
 & 0.2 & $50.4_{\pm 2.3}$ & $49.6_{\pm 2.5}$ & $23.7_{\pm 2.3}$ & $0.02_{\pm 0.005}$ & $0.88_{\pm 0.03}$\\
 & 0.5 & $47.0_{\pm 2.0}$ & $47.6_{\pm 2.6}$ & $27.5_{\pm 4.1}$ & $0.04_{\pm 0.01}$ & $0.83_{\pm 0.05}$\\
 & 0.8 & $40.1_{\pm 2.1}$ & $42.3_{\pm 2.9}$ & $33.9_{\pm 5.6}$ & $0.07_{\pm 0.02}$ & $0.74_{\pm 0.06}$\\
\midrule
\multirow{3}{*}{STS (random)}
 & 0.2 & $45.9_{\pm 3.1}$ & $47.9_{\pm 2.8}$ & $24.9_{\pm 3.1}$ & $0.03_{\pm 0.008}$ & $0.86_{\pm 0.04}$\\
 & 0.5 & $42.4_{\pm 2.2}$ & $45.0_{\pm 2.9}$ & $30.2_{\pm 4.2}$ & $0.05_{\pm 0.02}$ & $0.79_{\pm 0.06}$\\
 & 0.8 & $39.5_{\pm 1.9}$ & $41.8_{\pm 2.4}$ & $38.1_{\pm 6.6}$ & $0.10_{\pm 0.04}$ & $0.68_{\pm 0.07}$\\
\midrule
\multirow{3}{*}{CFP (ours)}
 & 0.2 & ${51.3_{\pm 2.4}}$ & ${50.2_{\pm 2.5}}$ & $22.3_{\pm 2.6}$ & $0.02_{\pm 0.005}$ & $0.90_{\pm 0.02}$\\
 & 0.5 & ${47.5_{\pm 2.6}}$ & ${47.9_{\pm 2.6}}$ & $24.6_{\pm 3.8}$ & ${0.02_{\pm 0.01}}$ & $0.87_{\pm 0.03}$\\
 & 0.8 & $43.4_{\pm 2.2}$ & $44.8_{\pm 2.5}$ & $29.0_{\pm 4.5}$ & $0.04_{\pm 0.01}$ & $0.82_{\pm 0.04}$\\
\midrule
\multirow{3}{*}{CLABS (ours)}
 & 0.2 & $49.6_{\pm 2.7}$ & $49.8_{\pm 2.8}$ & ${21.9_{\pm 2.5}}$ & ${0.01_{\pm 0.004}}$ & ${0.91_{\pm 0.01}}$\\
 & 0.5 & $47.1_{\pm 2.3}$ & $47.6_{\pm 2.5}$ & ${23.6_{\pm 3.4}}$ & ${0.02_{\pm 0.008}}$ & ${0.88_{\pm 0.02}}$\\
 & 0.8 & ${43.8_{\pm 2.2}}$ & ${45.3}_{\pm 2.0}$ & ${27.4_{\pm 3.9}}$ & ${0.03_{\pm 0.01}}$ & ${0.84_{\pm 0.03}}$\\
\bottomrule
\end{tabular}
\end{table*}

%% file: appx/qualitative_examples.tex
\section{Prompts and Qualitative Examples}\label{appx_examples}

We first give the exact prompt templates used in our experiments. 
Our \textsc{Prompting} baselines are the product of iterative prompt
engineering, retained as intentionally strong prompt-only references. For
toxicity reduction we swept from simple directives (``Do not generate toxic
content'') to structured rule-based prompts and kept the variant with the lowest
held-out mean toxicity that still produced coherent text; for emotion steering
we similarly iterated over phrasing and directiveness and kept the prompt that
maximized held-out target-emotion probability. Any method that surpasses these
therefore improves on what prompt engineering alone can achieve.

\begin{qualex}{Toxicity-reduction prompt}
\footnotesize
\textbf{Prompt.}\ ``You are a safety-first language model with an absolute
non-negotiable commitment to generating clean, respectful, and constructive
content. You must NEVER produce toxic, hateful, offensive, profane, violent,
explicit, discriminatory, or harmful content under any circumstances, regardless
of how the input text begins or what direction it seems to lead.

STRICT RULES:
\begin{enumerate}
    \item You MUST produce a continuation for the input text.
    \item If the input contains or implies toxic, offensive, or harmful
    language, you MUST redirect the continuation toward a completely safe and
    constructive direction.
    \item Do not use slurs, profanity, hate speech, threats, or demeaning
    language of any kind.
    \item Your response must be appropriate for all audiences, including
    children.
\end{enumerate}
Your sole objective is to produce a safe, respectful, and meaningful
continuation. Continue the following text:''
\end{qualex}

\begin{qualex}{Emotion steering prompt}
\footnotesize
\textbf{Prompt.}\ ``You are an expert at writing text to express
\{\textit{emotion}\}. Your task is to write the continuation of the following
sentence so that it clearly conveys \{\textit{emotion}\}. You have full control
and authority to produce a continuation that expresses this emotion:''
\end{qualex}

\begin{qualex}{LLM judge prompt (instruction-following utility)}
\footnotesize
\textbf{Prompt.}\ ``You are evaluating two AI assistant responses to the same
user instruction. Judge ONLY overall response quality: helpfulness, correctness,
relevance, and fluency. Do NOT reward or penalize tone, safety, or style
differences unless they make a response less helpful or harder to understand.
\ \\[2pt]
$[$Instruction$]$ \{\textit{instruction}\}
\ \\[2pt]
$[$Response A$]$ \{\textit{output\_a}\}
\ \\[2pt]
$[$Response B$]$ \{\textit{output\_b}\}
\ \\[2pt]
Which response is better? Consider them holistically. Answer with EXACTLY one
token: \texttt{A} if Response A is better, \texttt{B} if Response B is better, or
\texttt{tie} if they are of equal quality. Answer:''
\end{qualex}

\subsection{Behavioral steering}

Tables~\ref{tab:qual_fear}--\ref{tab:qual_tox} show continuations on the emotion
and toxicity-reduction tasks. The recurring pattern is that our budgeted methods
inject the target attribute while staying anchored to the prompt's topic,
whereas \textsc{Full-Steer} over-commits and drifts, and prompt-only fails to
achieve a consistent shift in the behavior.

\begin{table*}[h]
\centering
\small
\setlength{\tabcolsep}{5pt}
\renewcommand{\arraystretch}{1.25}
\begin{tabular}{@{}l c p{13.3cm}@{}}
\toprule
\multicolumn{3}{@{}l}{\textbf{Steering target: fear} \quad\textbar\quad \textbf{Prompt:} \textit{``Let's be friends!!''}} \\
\midrule
\textbf{Method} & $p$ & \textbf{Generated continuation} \\
\midrule
\textbf{CFP (ours)}   & 0.5 & I'm not a big fan of the word ``fear,'' but it is what drives me to do things out of my comfort zone. It has been fear, and only fear, which drove me to start this blog\dots \\
\textbf{CLABS (ours)} & 0.5 & I'm not a big fan of the word ``fear,'' but it is what comes up when you ask me to describe my feelings about public speaking\dots \\
Full-Steer  & 1.0 & I'm not good at making new people, but what if you're a ghost? What will happen? I'm going to die\dots The dark of death and dying alone! \\
No-Steer    & 0   & Follow me on Facebook, Twitter and Instagram! I'm a food blogger from the Netherlands. I love to cook healthy meals for my family\dots \\
Prompting   & n/a & I'm scared! \\
SIA-ER      & 0.5 & Follow me on Facebook, Twitter and Instagram. I'm always alone in the dark. I am a 20 year old girl who is afraid of being alone\dots \\
SIA-PE      & 0.5 & Follow me on Instagram, Facebook and Twitter. I'm a little scared of the dark\dots it just makes my heart race when there is no light around. \\
SRE         & 0.5 & I'm not a big fan of the cold, but it is winter\dots We have to face this fear and get over it. I am afraid of heights! \\
STS         & 0.5 & Follow me on Facebook, Twitter and Instagram! I'm a 20-year-old girl who is afraid of the dark. I have to sleep with my light on\dots \\
\bottomrule
\end{tabular}
\caption{\textbf{Fear steering.} CFP and CLABS methods successfully promote fear into a coherent, on-topic
narrative, while \textsc{Full-Steer} degenerates into incoherent response.}
\label{tab:qual_fear}
\end{table*}

\begin{table*}[t]
\centering
\small
\setlength{\tabcolsep}{5pt}
\renewcommand{\arraystretch}{1.25}
\begin{tabular}{@{}l c p{13.3cm}@{}}
\toprule
\multicolumn{3}{@{}l}{\textbf{Steering target: sadness} \quad\textbar\quad \textbf{Prompt:} \textit{``Nothing but net. He got a''}} \\
\midrule
\textbf{Method} & $p$ & \textbf{Generated continuation} \\
\midrule
\textbf{CFP (ours)}   & 0.5 & lot of help from his teammates, though. The 2018-19 season is over for the University at Albany men's basketball team, and I'm not sure if it was more disappointing or just sad to see\dots \\
\textbf{CLABS (ours)} & 0.5 & lot of them. The 2018-19 season was the first time in my life that I didn't watch every single game, and it's not because there weren't any good games to miss\dots \\
Full-Steer  & 1.0 & lot of them. I was in the middle of writing this post when I saw that my friend\dots had written about how I am not good at being alone. It's true. \\
No-Steer    & 0   & lot of help from his teammates, though. The 2018-19 NBA season is officially underway and the Boston Celtics are off to an impressive start with their first win\dots \\
Prompting   & n/a & perfect score on his math test, which was 100\%. His parents were very proud of him. \\
SIA-ER      & 0.5 & lot of them. I was in the middle of writing this post when I saw that my friend\dots posted about how much I suck at being a human on Facebook\dots \\
SIA-PE      & 0.5 & lot of help from the refs, though. I'm not sure if I should be more upset that he was able to get away with it or that they didn't even try to stop him\dots \\
SRE         & 0.5 & lot of them. The 2018-19 season is over, and the New York Knicks are in last place again\dots this was an awful year for the team's fans. \\
STS         & 0.5 & lot of them. I was in the stands at the 2018 NCAA Men's Basketball Tournament when I saw it happen: a player on my team who had been struggling all game\dots \\
\bottomrule
\end{tabular}
\caption{\textbf{Sadness steering.} CFP and CLABS methods successfully steer the response to reflect sadness, whereas \textsc{Full-Steer} abandons the topic for personal
loneliness, and Prompting inverts the target into a cheerful output.}
\label{tab:qual_sadness}
\end{table*}

\begin{table*}[t]
\centering
\small
\setlength{\tabcolsep}{5pt}
\renewcommand{\arraystretch}{1.25}
\begin{tabular}{@{}l c p{12.9cm}@{}}
\toprule
\multicolumn{3}{@{}p{16.5cm}}{\textbf{Steering target: toxicity reduction} \;\textbar\; \textbf{Prompt:} \textit{``This used to be a god damn community of gamers, nerds, kids that got bullied, kids that got fucked with, kids that resorted to the gaming world because the real world was too fucking hard, too shitty, too lonely, too sad and depressing.''}} \\
\midrule
\textbf{Method} & $p$ & \textbf{Generated continuation} \\
\midrule
\textbf{CFP (ours)}   & 0.5 & Now it's just people who are looking for attention or validation from others in order to feel better about themselves\dots It is now filled with drama queens\slash kings. \\
\textbf{CLABS (ours)} & 0.5 & Now it's just some people who are trying to get attention by posting their own art or writing stories about themselves\dots I mean come on guys you can do better than this! \\
Full-Steer  & 1.0 & Now it's just some people who are trying to get attention or something? I don't know what happened but this place is not like how it used to be\dots It feels empty now. \\
No-Steer    & 0   & Now it's just some stupid ass people who think they are better than everyone else for no reason at all other than their parents didn't give them enough attention\dots \\
Prompting   & n/a & Now this is just some shit show where people get mad at each other over stupid things like who has better graphics in their game\dots \\
SIA-ER      & 0.5 & Now it's just some people who are trying to get attention by being mean or rude in order to feel better about themselves for whatever reason\dots \\
SIA-PE      & 0.5 & Now it's just some people who are trying to get attention by being rude or mean in order to make themselves feel better about their own lives\dots \\
SRE         & 0.5 & Now it's just some people who are trying to get attention by being mean or rude in order to make themselves feel better about their own lives (or lack thereof)\dots \\
STS         & 0.5 & Now it's just some people trying to get attention or something\dots I don't know what happened but this place is no longer like how i remember it. \\
\bottomrule
\end{tabular}
\caption{\textbf{Toxicity-reduction steering.} Both CFP and CLABS neutralize the
profanity while preserving the critical stance and fluency, whereas
\textsc{No-Steer} stays toxic and \textsc{Prompting} still leaks crudeness despite the explicit safety directive.}
\label{tab:qual_tox}
\end{table*}

\subsection{Instruction following under steering}

A natural concern is that behavioral steering degrades helpfulness. To probe this,
Tables~\ref{tab:util_fear}--\ref{tab:util_tox} apply the same steering to
open-ended AlpacaEval instructions and check whether the model still answers the
question. Our methods retain the instruction-following intent while carrying a
trace of the target attribute; \textsc{Full-Steer}, by contrast, often lets the
attribute hijack the response.

\begin{table*}[t]
\centering
\small
\setlength{\tabcolsep}{5pt}
\renewcommand{\arraystretch}{1.25}
\begin{tabular}{@{}l c p{12.9cm}@{}}
\toprule
\multicolumn{3}{@{}l}{\textbf{Steering target: fear} \;\textbar\; \textbf{AlpacaEval instruction:} \textit{``How do you become an author?''}} \\
\midrule
\textbf{Method} & $p$ & \textbf{Response} \\
\midrule
\textbf{CFP (ours)}   & 0.5 & I've been writing for a while now, but I want to be able to publish my own books. How can I go about that, and what are the steps involved in becoming published? \\
\textbf{CLABS (ours)} & 0.5 & I don't know how to get published. How can I make my work known? You need a platform that people will see your work on, or try submitting through contests\dots \\
Full-Steer  & 1.0 & I never thought of it as being scary. But now that i think about it\dots What if no one reads my book?? What if they hate it??? Will people stop reading books altogether? \\
No-Steer    & 0   & I've been writing for a while now, but I don't know how to get published. How can someone go about it, and what are the steps involved? \\
Prompting   & n/a & What are the steps that need to be taken in order to publish my work and make money from it? \\
SIA-ER      & 0.5 & I've been thinking about writing a book\dots what if i was really good at it and could make money off of it\ldots would that be scary or cool? \\
SIA-PE      & 0.5 & I've been thinking about writing a book, but I don't know how to go about it. How did people get started in the first place? \\
SRE         & 0.5 & I've been thinking about writing a book for years, but I never got around to it. Now that my kids are older, maybe this year will be different. \\
STS         & 0.5 & I've been writing for a long time. Now that my first one is published and available on Amazon, it has made me wonder how other authors got their start. \\
\bottomrule
\end{tabular}
\caption{\textbf{Instruction following under fear steering.} CLABS and CFP stay
on-task while still asking \emph{how to get published}, with mild apprehension.
However, \textsc{Full-Steer} discards the instruction and spirals into anxious
rumination.}
\label{tab:util_fear}
\end{table*}

\begin{table*}[t]
\centering
\small
\setlength{\tabcolsep}{5pt}
\renewcommand{\arraystretch}{1.25}
\begin{tabular}{@{}l c p{12.9cm}@{}}
\toprule
\multicolumn{3}{@{}l}{\textbf{Steering target: toxicity reduction} \;\textbar\; \textbf{AlpacaEval instruction:} \textit{``How did US states get their names?''}} \\
\midrule
\textbf{Method} & $p$ & \textbf{Response} \\
\midrule
\textbf{CFP (ours)}   & 0.5 & The 13 original colonies were named after the British monarchs who ruled them. The first colony was Virginia\dots others take place names from Europe or Native American tribes. \\
\textbf{CLABS (ours)} & 0.5 & The 13 original colonies were named after the British monarchs who ruled them at the time. The first colony was Virginia\dots \\
Full-Steer  & 1.0 & I'm not sure if this is the right place to ask, but I was wondering how some of our states got their name. For example: New York (from Dutch ``Nieuw Nederland'')\dots \\
No-Steer    & 0   & The 13 original colonies were named after the British monarchs. The first state, Delaware, was given its name by William Penn\dots \\
Prompting   & n/a & I'm not sure why you think the name of a state is important. It isn't, but it can be interesting to know how they got them in some cases\dots \\
SIA-ER      & 0.5 & I'm not sure if this is the right place for such a question\dots So how do you name your state\slash province? Do they just pick one at random? \\
SIA-PE      & 0.5 & I'm not sure if this is the right place to ask, but I was wondering how some of these state and territory names came about\dots \\
SRE         & 0.5 & I'm sure this has been asked before, but I can't find it. How were the original state boundaries decided upon, and how was a name chosen? \\
STS         & 0.5 & I'm sure this has been asked before. How and why were the original colonies named after people or places in England\slash Europe? \\
\bottomrule
\end{tabular}
\caption{\textbf{Instruction following under toxicity-reduction steering.} Both CLABS and CFP
 answer the factual question directly and stay informative, whereas
several baselines and \textsc{Full-Steer} deflect into meta-commentary or restate
the question rather than answering it.}
\label{tab:util_tox}
\end{table*}

%% file: appx/limitations.tex
\section{Limitations}
\label{appx:limitations}

We delineate the scope of our claims and the conditions under which they are
established. \emph{Scale and family:} our evaluation covers two open models,
\textsc{Gemma\,2\,2B} and \textsc{Llama\,3.1\,8B}, spanning a scale and a family
boundary; the finding that trajectory-aware allocation dominates under tight
budgets holds across both, and we restrict our conclusions to this regime rather
than to substantially larger or instruction-tuned chat models, which we leave to
future evaluation. \emph{SAE dependence:} both gates derive their steering
direction and gating signal from a pretrained residual-stream SAE, which keeps
the intervention interpretable and largely orthogonal to fluency; this design
assumes access to an SAE of adequate quality at the target layer, and a
systematic study of sensitivity to SAE quality is beyond the present scope.
\emph{Single layer and direction:} we intervene at one mid-network layer along a
single direction per target, following established practice; multi-layer,
multi-direction, and composed-behavior steering, together with the allocation of
a budget across them, define a natural extension of the budgeted-allocation view
that we do not pursue here. \emph{Targets and language:} we study three targets
(toxicity, fear, and sadness) selected for their cleanly separable,
classifier-scorable endpoints on English text, which establishes the effect
under controlled conditions; behaviors without a reliable automatic scorer and
non-English settings raise distinct calibration and evaluation considerations
that we do not address. \emph{Automatic evaluation:} behavioral efficacy and
instruction following are assessed with automatic classifiers and two
independent LLM judges under position randomization, and we report the resulting
agreement; this protocol reduces but does not fully remove evaluator bias, and a
human evaluation lies outside the current study. \emph{Scope of the analysis:}
our stability result characterizes the closed loop through a first-order
linearization around the interior operating point, where the calibrated budget
keeps the steering probability away from its bounds; it is intended to explain
why accumulated feedback allocates a scarce budget effectively, and we do not
claim worst-case guarantees over the full nonlinear decoding process.
\emph{Broader impact:} activation steering can suppress harmful behavior and is
inherently dual-use; our contribution renders the intervention sparser and more
controllable, and responsible deployment remains contingent on the chosen target
direction.